\PassOptionsToPackage{prologue,dvipsnames}{xcolor}
\documentclass[lettersize,journal]{IEEEtran}
\usepackage{colortbl}
\usepackage{amsmath,amsfonts}
\usepackage{algorithmic}
\usepackage{algorithm}
\usepackage{array}
\usepackage[caption=false,font=normalsize,labelfont=sf,textfont=sf]{subfig}
\usepackage[table]{xcolor}
\usepackage{multirow}
\usepackage{url}
\usepackage{graphicx}
\usepackage{diagbox}
\usepackage{cite}
\usepackage {makecell}
\usepackage{multirow}
\usepackage[pagewise]{lineno}
\usepackage[dvipsnames]{xcolor}

\usepackage{ulem}
\usepackage{booktabs}
\newcommand\clr[1]{{\color{black}{#1}}}
\usepackage[breaklinks]{hyperref}
\usepackage{ulem}
\usepackage{url}

\usepackage{scalerel}
\usepackage{color}
\usepackage{tikz}
\usetikzlibrary{svg.path}
\definecolor{orcidlogocol}{HTML}{A6CE39}
\tikzset{
  orcidlogo/.pic={
    \fill[orcidlogocol] svg{M256,128c0,70.7-57.3,128-128,128C57.3,256,0,198.7,0,128C0,57.3,57.3,0,128,0C198.7,0,256,57.3,256,128z};
    \fill[white] svg{M86.3,186.2H70.9V79.1h15.4v48.4V186.2z}
                 svg{M108.9,79.1h41.6c39.6,0,57,28.3,57,53.6c0,27.5-21.5,53.6-56.8,53.6h-41.8V79.1z M124.3,172.4h24.5c34.9,0,42.9-26.5,42.9-39.7c0-21.5-13.7-39.7-43.7-39.7h-23.7V172.4z}
                 svg{M88.7,56.8c0,5.5-4.5,10.1-10.1,10.1c-5.6,0-10.1-4.6-10.1-10.1c0-5.6,4.5-10.1,10.1-10.1C84.2,46.7,88.7,51.3,88.7,56.8z};
  }
}
\newcommand\orcidicon[1]{\href{https://orcid.org/#1}{\mbox{\scalerel*{
\begin{tikzpicture}[yscale=-1,transform shape]
\pic{orcidlogo};
\end{tikzpicture}
}{|}}}}
\usepackage{hyperref}
\hypersetup{
    colorlinks=true,
    linkcolor=blue,
    filecolor=black,      
    urlcolor=magenta,
    citecolor=purple
}
\usepackage{tikz-network}
\definecolor{cvprblue}{rgb}{0.21,0.49,0.74}
\newcommand{\settablefont}{\fontsize{6.5}{10.1}\selectfont}
\newcolumntype{C}[1]{>{\centering\arraybackslash}p{#1}}
\newcommand\eg{\textit{e.g}}
\newcommand\ie{\textit{i.e}}

\title{LIX: Implicitly Infusing Spatial Geometric\\Prior Knowledge into Visual Semantic Segmentation\\for Autonomous Driving
}
\author{
Sicen Guo$^{\orcidicon{0009-0000-8079-8056}\,}$, Ziwei Long$^{\orcidicon{0009-0007-1520-5526}\,}$, Zhiyuan Wu$^{\orcidicon{0009-0001-7253-7603}\,}$, Qijun Chen, ~\IEEEmembership{Senior Member, ~IEEE},\\ \ \ \ Ioannis Pitas, ~\IEEEmembership{Life Fellow, ~IEEE}, Rui Fan$^{\orcidicon{0000-0003-2593-6596}\,}$, ~\IEEEmembership{{Senior} Member, ~IEEE}
\thanks{(\emph{Corresponding author: Rui Fan})}
\thanks{Sicen Guo, Ziwei Long, Zhiyuan Wu, and Qijun Chen are with the College of Electronics \& Information Engineering, Tongji University, Shanghai 201804, China (e-mail: \{guosicen, zwlong, gwu, qjchen\}@tongji.edu.cn)}
\thanks{Ioannis Pitas is with the Department of Informatics, University of Thessaloniki, 541 24 Thessaloniki, Greece (e-mail: pitas@csd.auth.gr).}
\thanks{Rui Fan is with the College of Electronics \& Information Engineering, Shanghai Institute of Intelligent Science and Technology, Shanghai Research Institute for Intelligent Autonomous Systems, the State Key Laboratory of Intelligent Autonomous Systems, and Frontiers Science Center for Intelligent Autonomous Systems, Tongji University, Shanghai 201804, P. R. China, as well as with the National Key Laboratory of Human-Machine Hybrid Augmented Intelligence, Xi'an Jiaotong University, Xi'An, Shaanxi 710049, P. R. China (e-mail: rui.fan@ieee.org).}
}

\begin{document}


\maketitle
\normalem
\begin{abstract}
Despite the impressive performance achieved by data-fusion networks with duplex encoders for visual semantic segmentation, they become ineffective when spatial geometric data are not available. \clr{Implicitly infusing the spatial geometric prior knowledge acquired by a data-fusion teacher network into a single-modal student network is a practical, albeit less explored research avenue.} This article delves into this topic and resorts to knowledge distillation approaches to address this problem. We introduce the Learning to Infuse ``X'' (LIX) framework, with novel contributions in both logit distillation and feature distillation aspects. We present a mathematical proof that underscores the limitation of using a single, fixed weight in decoupled knowledge distillation and introduce a logit-wise dynamic weight controller as a solution to this issue. Furthermore, we develop an adaptively-recalibrated feature distillation algorithm, including two novel techniques: feature recalibration via kernel regression and in-depth feature consistency quantification via centered kernel alignment. Extensive experiments conducted with intermediate-fusion and late-fusion networks across various public datasets provide both quantitative and qualitative evaluations, demonstrating the superior performance of our LIX framework when compared to other state-of-the-art approaches. 
\end{abstract}


\section{Introduction}
\label{sec:intro}
\IEEEPARstart{I}{n} the domain of data-driven autonomous driving perception, a prevailing consensus among researchers asserts that ``the increased availability of well-annotated training data is strongly correlated with improved learning performance.'' \clr{When examining visual semantic segmentation as an illustrative case, the adoption of data-fusion networks, equipped with duplex encoders to acquire knowledge from both RGB images and spatial geometric information, consistently demonstrate superior performance compared to conventional single-modal networks trained solely on RGB images \cite{li2024roadformer,fan2020sne-roadseg, zhang2023cmx, yin2023dformer}. This performance improvement is due to their ability to learn heterogeneous features from diverse data sources \cite{zhang2023delivering}. RGB images primarily capture rich color and texture information, whereas other visual data sources, commonly designated as ``X'',  contain informative spatial geometric information.} The fusion of their features allows for a more comprehensive understanding of the driving environment.

\clr{However, a significant limitation of data-fusion networks stems from their dependence on the availability of the ``X'' data, which can pose constraints in scenarios devoid of range sensors \cite{wu2024s}. Additionally, when the accuracy of the ``X'' data falls below expectations, possibly due to issues such as variations in camera-LiDAR calibration, the fusion of these heterogeneous features can potentially lead to a degradation in the overall performance of visual semantic segmentation \cite{li2024roadformer}. As a result, the implicit infusion of spatial geometric prior knowledge from a teacher network (a data-fusion network trained with RGB-X data) to a student network (a single-modal network trained exclusively with RGB images) emerges as an interesting research direction worth pursuing. It is reasonable to consider that knowledge distillation (KD) techniques can be a viable solution to achieve this objective.}

\clr{Existing KD techniques are generally classified into two main categories: logit distillation (LD) and feature distillation (FD). The former approaches \cite{10292885, shu2021channel, hinton2015distilling} train the student network to replicate the logits of the teacher network by minimizing the divergence or distance between the probability distributions generated by both networks. On the other hand, the latter approaches \cite{zagoruyko2016paying, kim2018paraphrasing, huang2017like, park2019relational, tung2019similarity, ahn2019variational, chen2021distilling} leverage the abundant information available in the activations, neurons, and features of the teacher network's intermediate layers to provide guidance and supervision for the training of the student network.}

Nonetheless, directly applying these existing techniques to our specific problem remains challenging due to three considerations. \clr{First, while there are several existing KD methods for cross-modal distillation, such as \cite{zheng2022boosting1, zhou2023unidistill, cen2023cmdfusion} (from RGB images and LiDAR point clouds to LiDAR point clouds) and \cite{zheng2022boosting2,qiu2023multi} (from multi-frame point clouds to single-frame point clouds), they are not specifically designed for the task discussed in this study. Direct application of such algorithms to our task is infeasible due to the differences in network architectures that arise from distinct input modalities. Second, differences in architecture between teacher and student networks result in discrepancies in heterogeneous feature characteristics, such as dimensions, magnitudes, and distributions. These discrepancies form a barrier to the effective infusion of spatial geometric prior knowledge \cite{aggarwal2015data}. Third, comprehensive feature consistency measurement should be emphasized and requires further exploration, as it directly impacts the overall performance of FD \cite{huang2017like,zagoruyko2016paying,park2019relational}.}

This article introduces \uline{\textbf{L}earning to \textbf{I}nfuse ``\textbf{X}'' (\textbf{LIX})} framework (see Fig. \ref{fig.framework}), designed to implicitly infuse spatial geometric prior knowledge acquired from a data-fusion teacher network into a single-modal student network. We make three major contributions to address the above-mentioned limitations. We begin by revisiting the LD theory based on decoupled knowledge distillation (DKD) \cite{zhao2022decoupled} and reformulate its loss as a weighted combination of target class LD (TCLD) and non-target class LD (NCLD) losses. By deriving the gradient of the LD loss function with respect to the student logit, we expose the limitations of the DKD algorithm, which relies on a single, fixed weight. Consequently, we design a dynamic weight controller (DWC), capable of generating a weight for each logit, thereby improving the overall LD performance. As for FD, we first introduce an adaptive feature recalibration approach based on kernel regression, which aligns the features of teacher and student networks across various dimensions (spatial, channel, magnitude, and distribution). Finally, we resort to the centered kernel alignment (CKA) \cite{nguyen2020wide} algorithm based upon Hilbert-Schmidt independence criterion (HSIC) \cite{ma2020hsic} to formulate our novel FD loss, which quantifies the feature consistency between teacher and student networks. These contributions collectively improve the effectiveness of implicitly infusing spatial geometric prior knowledge into visual semantic segmentation for autonomous driving. In a nutshell, our contributions can be summarized as follows:
\begin{itemize}
    \item We implicitly infuse spatial geometric prior knowledge into visual semantic segmentation by distilling an RGB-X data-fusion teacher network into a single-modal student network that operates solely on RGB images.
    \item We present the novel dynamically-weighted LD (DWLD) algorithm, which extends the DKD algorithm, by assigning an appropriate weight to each logit, resulting in better performance compared to the baseline algorithm. 
    \item We introduce the novel adaptively-recalibrated FD (ARFD) algorithm that performs feature recalibration via kernel regression and feature consistency measurement leveraging HSIC-based CKA.
    \item We have conducted extensive experiments using representative RGB-X semantic segmentation networks on multiple public datasets to quantitatively and qualitatively validate the effectiveness of our introduced novel LD and FD techniques. 
\end{itemize}
The remainder of this article is structured as follows: Sect. \ref{sec:related} reviews related works. Sect. \ref{sec:methodology} details our proposed LIX. Sect. \ref{sec:experiments} compares LIX with other SoTA methods and presents the ablation study results. Limitations and extendability of our proposed LIX framework are discussed in Sect. \ref{sec:discussion}. Finally, in Sect. \ref{sec:conclusions}, we summarize this article and provide recommendations for future work.

\section{Related Work}
\label{sec:related}

\subsection{RGB-X Semantic Segmentation}
\label{sec:rgb-x}
According to the data-fusion stage, state-of-the-art (SoTA) RGB-X semantic segmentation networks can be grouped into three classes: early-fusion, intermediate-fusion, and late-fusion networks \cite{zhang2021deep}. Early-fusion approaches generally combine RGB and X images at the input level. Such a straightforward yet simplistic data-fusion strategy has limitations in capturing a deep understanding of the environment \cite{zhang2021deep}. In contrast, intermediate-fusion approaches \cite{fan2020sne-roadseg,wang2021sne} typically extract heterogeneous features from RGB and X images using duplex encoders. These features are subsequently fused within the encoder to fully exploit their inherent characteristics. Similar to intermediate-fusion methods, late-fusion approaches \cite{ha2017mfnet,valada2017adapnet,cheng2017locality} use two parallel encoders (one for RGB images and one for X data) to extract heterogeneous features. However, these methods primarily focus on data fusion within the decoder. In this article, we utilize two representative data-fusion models as our baseline networks, namely SNE-RoadSeg \cite{fan2020sne-roadseg} (a computationally intensive, intermediate-fusion network) and MFNet \cite{ha2017mfnet} (a lightweight, late-fusion network), to comprehensively validate the effectiveness of our proposed LIX framework. 

\subsection{Knowledge Distillation}
\label{sec.knowledge_distillation}
KD methods have demonstrated significant potential in balancing accuracy and efficiency across various applications. In the domain of object detection, instance-aware distillation (InsDist) \cite{10024393} enhances feature learning by decoupling instance-related features and modeling inter-instance relationships. Moreover, smoothed teacher (ST) \cite{10292885} focuses on reducing intra-class variance to improve the student network's performance. Additionally, RadarDistill \cite{bang2024radardistill} refines Radar representations by transferring knowledge acquired from LiDAR point clouds through FD. In the field of image segmentation, MSTNet-KD \cite{zhou2024mstnet} enables a compact student network to learn semantic information extracted from a complex teacher network based on a multi-level semantic alignment technique. These innovations highlight KD's versatility across various modalities and tasks, underscoring its broad applicability in real-world scenarios.

The first LD method \cite{hinton2015distilling} guides the student network to mimic the distribution characteristics of soft targets, which are generated by the teacher network). Unlike the original KD \cite{hinton2015distilling}, channel-wise KD (CWKD) \cite{shu2021channel} algorithm normalizes the activations within each channel to generate a probability map. Subsequently, the Kullback-Leibler (KL) divergence \cite{hinton2015distilling} is utilized to minimize the discrepancy between the probability maps generated by teacher and student networks. 
Moreover, weighted soft label distillation (WSLD) \cite{zhourethinking} adjusts the sample-wise bias-variance trade-off during model training by re-weighting the soft label loss. However, in these previous works \cite{hinton2015distilling, shu2021channel, zhourethinking, huang2022knowledge}, the TCLD and NCLD terms are treated as interdependent and coupled terms. The recent study DKD \cite{zhao2022decoupled} suggests that the coupled formulation of KD limits the effectiveness and flexibility of knowledge transfer, and it is necessary to reformulate the classical KD loss into two independent and decoupled terms. 

LD methods depend exclusively on output logits and do not incorporate supervision to ensure feature consistency between teacher and student networks. Such feature-level supervision has been demonstrated to be crucial for effective representation learning \cite{liu2015representation}. The first FD method, FitNet \cite{romero2014FitNets}, directly matches feature activations between teacher and student networks. Attention transfer (AT) \cite{zagoruyko2016paying} resorts to both activation-based and gradient-based spatial attention maps to realize FD. Several subsequent studies, \eg, similarity-preserving (SP) KD \cite{tung2019similarity} and variational information distillation (VID) \cite{ahn2019variational}, have sought to reduce the dimension of features to accommodate the increasing depth of networks. Factor transfer (FT) \cite{kim2018paraphrasing} introduces a paraphraser and a translator to extract and transfer structured knowledge, whereas relational knowledge distillation (RKD) \cite{park2019relational} measures structural consistency between teacher and student networks using distance-wise and angle-wise losses. Neuron selectivity transfer (NST) \cite{huang2017like} formulates KD as a distribution matching problem by aligning neuron selectivity patterns through maximum mean discrepancy minimization. More recent methods \cite{huangmasked,fan2023augmentation,zhang2020prime} enable the model to concentrate on the most informative parts of the data. For instance, masked distillation (MasKD) \cite{huangmasked} employs spatial masks to emphasize essential feature regions, while augmentation-free dynamic curriculum distillation (Af-DCD) \cite{fan2023augmentation} adjusts task difficulty based on the student's learning progress. Moreover, prime-aware adaptive distillation (PAD) \cite{zhang2020prime} prioritizes informative samples via uncertainty metrics, which is particularly beneficial for class-imbalanced datasets.

However, the aforementioned algorithms are not designed specifically to infuse spatial geometric prior knowledge for semantic segmentation. In this article, we make significant advancements in both LD and FD. We introduce a logit-wise dynamic weight controller to address the limitations of using a single, fixed weight in DKD. Additionally, we develop an adaptively-recalibrated feature distillation algorithm that includes feature recalibration and feature consistency quantification. Extensive experiments demonstrate the superior performance of our LIX framework over all algorithms reviewed.

\section{Methodology}
\label{sec:methodology}
\subsection{Overall Workflow}
\label{sec:workflow}
\begin{figure*}[t!]
\centering
\includegraphics[width=0.999\textwidth]{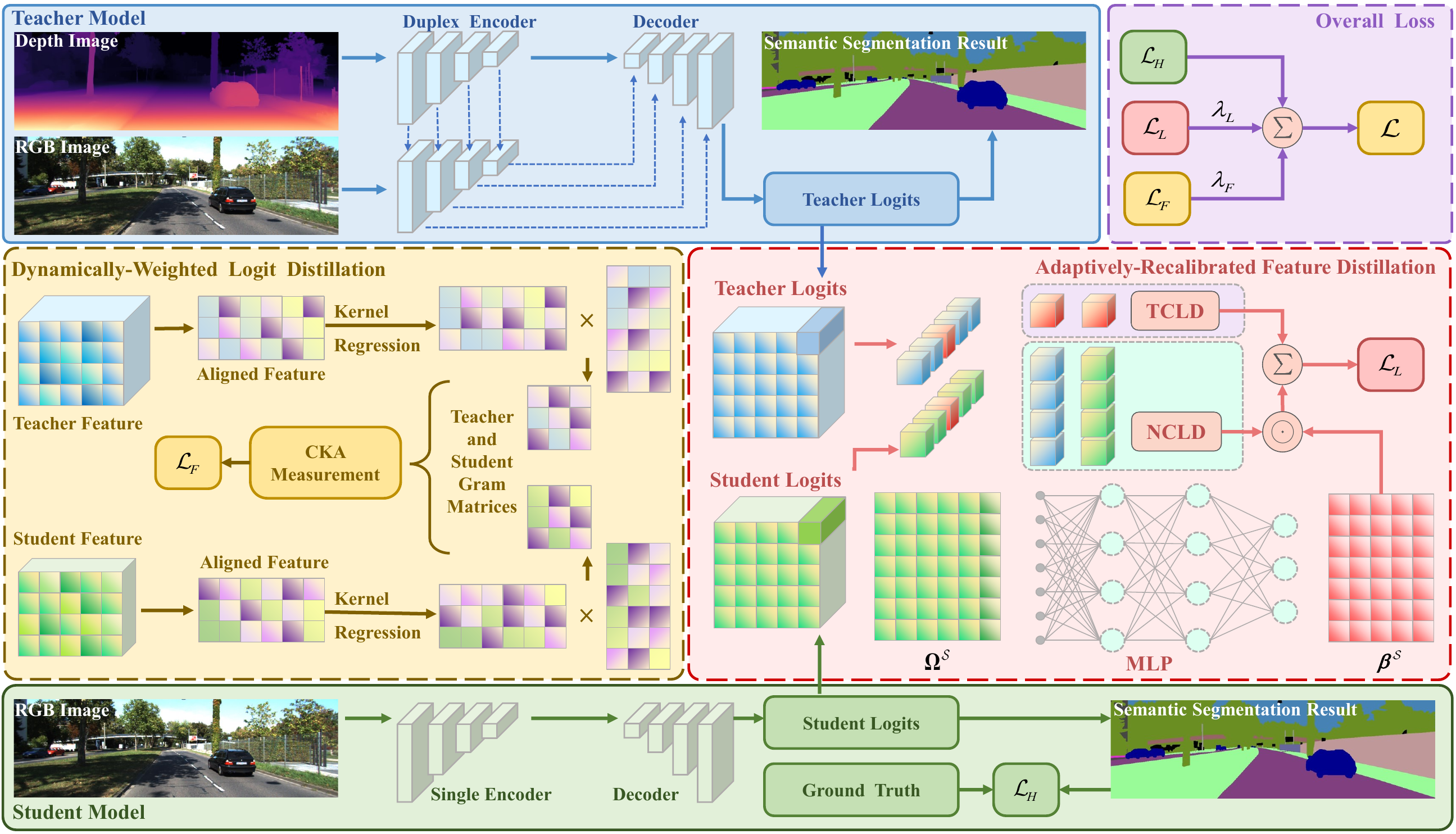}
\centering
\caption
{
An illustration of our proposed {\textbf{LIX framework}}, which consists of two key components: a) {\textbf{dynamically-weighted logit distillation}} and b) {\textbf{adaptively-recalibrated feature distillation}}.
}
\label{fig.framework}
\end{figure*}

As depicted in Fig. \ref{fig.framework}, our proposed LIX framework can implicitly infuse spatial geometric prior knowledge acquired from a data-fusion teacher network into a single-modal student network. DWLD first reformulates the LD loss as a weighted combination of TCLD and NCLD losses. Then, a dynamic weight controller generates a weight for each student logit and formulates the overall LD loss $\mathcal{L}_L$. As for FD, an adaptive feature recalibration approach based on kernel regression first aligns the features of teacher and student networks across various dimensions (spatial, channel, magnitude, and distribution). Subsequently, the CKA algorithm based upon HSIC quantifies the consistency between different features and formulates our FD loss $\mathcal{L}_F$. In summary, the overall loss can be formulated as a combination of the initial loss $\mathcal{L}_H$ of hard labels (ground truth) and distillation losses.

\subsection{Dynamically-Weighted Logit Distillation}
\label{sec:logit}

Most exiting LD methods \cite{hinton2015distilling,shu2021channel} primarily resorted to more effective regularization and optimization methods, rather than proposing novel distillation strategies. In this subsection, we delve deeper into the LD theory and reframe its loss as the weighted combination of TCLD and NCLD losses. We derive the gradient of the LD loss with respect to the student logit and expose limitations of the traditional DKD algorithm \cite{zhao2022decoupled}, which relies on a single, fixed weight. Drawing inspiration from these findings, we introduce a novel LD method, which is capable of dynamically generating a weight for each logit, thereby improving the overall performance of spatial geometric prior knowledge infusion.

\subsubsection{\textbf{Reformulating Logit Distillation}}
Let $\hat{\boldsymbol{p}} =
\left[ 
\hat{p}_1,
\dots,
\hat{p}_k,
\dots,
\hat{p}_C
\right]^\top \in \mathbb{R}^{C}$ be a column vector, storing probabilities $\hat{p}_i=P(y=i|\boldsymbol{q})={\exp \left(z_i\right)}/{\sum_{j=1}^C \exp \left(z_j\right)}$ ($\forall i\in[1,C]\cap\mathbb{Z}$) of pixel ${\boldsymbol{q}}$ belonging to $C$ classes, where $y$ denotes the predicted label of ${\boldsymbol{q}}$ and $z_i$ denotes its logit with respect to the $i$-th class. Let $\boldsymbol{b}=\left[ \hat{p}_k, 1-\hat{p}_k \right]^\top \in \mathbb{R}^{2}$ be a binary probability vector of the target class $k$. Let 
{
$
\hat{\boldsymbol{p}}_{{\backslash k}}=\hat{\boldsymbol{p}}/\left(1-\hat{p}_k\right)=
\left[ 
\hat{p}_{1,{\backslash k}},
\dots, 
\hat{p}_{k-1,{\backslash k}},
\hat{p}_{k+1,{\backslash k}},
\dots,
\hat{p}_{C,{\backslash k}},
\right]^\top\in \mathbb{R}^{(C-1)}
$
}
be a column vector, storing independently modeled probabilities among non-target classes (\ie, without considering the $k$-th class). The conventional LD uses the KL divergence
\cite{hinton2015distilling} between output probabilities $\hat{\boldsymbol{p}}^{\mathcal{T}}$ and $\hat{\boldsymbol{p}}^{\mathcal{S}}$ of the teacher and student networks, respectively. Its loss function is expressed as follows\footnote{$\mathcal{T}$ and $\mathcal{S}$ denote teacher and student networks, respectively.}: 
\begin{equation}
\begin{aligned}
\mathrm{LD} & 
=\mathrm{KL}
\big(
\hat{\boldsymbol{p}}^{\mathcal{T}} \| \hat{\boldsymbol{p}}^{\mathcal{S}}
\big) 
=\hat{p}^{\mathcal{T}}_k 
\log 
\frac{
\hat{p}^{\mathcal{T}}_k 
}
{
\hat{p}^{\mathcal{S}}_k 
}
+\sum_{i=1, i \neq k}^C 
\hat{p}^{\mathcal{T}}_i 
\log 
\frac{
\hat{p}^{\mathcal{T}}_i 
}
{
\hat{p}^{\mathcal{S}}_i 
}
,
\end{aligned}
\end{equation} 
which can be reformulated as follows \cite{zhao2022decoupled}:
\begin{equation}
\begin{aligned}
\mathrm{LD}  
& =\underbrace{\hat{p}^{\mathcal{T}}_k 
\log \frac{\hat{p}^{\mathcal{T}}_k }{\hat{p}^{\mathcal{S}}_k }
+\left(1-\hat{p}^{\mathcal{T}}_k\right) 
\log \frac{1-\hat{p}^{\mathcal{T}}_k}{1-\hat{p}^{\mathcal{S}}_k}}
_{\mathrm{KL}\left(\boldsymbol{b}^{\mathcal{T}} \| \boldsymbol{b}^{\mathcal{S}}\right)}
\\&+\left(1-\hat{p}^{\mathcal{T}}_k\right) 
\underbrace{\sum_{i=1, i \neq k}^C {\hat{p}^{\mathcal{T}}}_{{i,\backslash k}} 
\log \frac{{\hat{p}^{\mathcal{T}}}_{{i,\backslash k}}}{{\hat{p}^{\mathcal{S}}}_{{i,\backslash k}}}}
_{\operatorname{KL}\left({\hat{\boldsymbol{p}}^{\mathcal{T}}}_{{\backslash k}} \| {\hat{\boldsymbol{p}}^{\mathcal{S}}}_{{\backslash k}}\right)} . \ \ \ \  \ \ \ \  \ \ \ \  \ \ \ \
\label{eq:KD2}
\end{aligned}
\end{equation}
(\ref{eq:KD2}) can be rewritten as follows:
\begin{equation}
\begin{aligned}
\mathrm{LD}&=\underbrace{\mathrm{KL}\left(\boldsymbol{b}^{\mathcal{T}} \| \boldsymbol{b}^{\mathcal{S}}\right)}_{\mathrm{TCLD}}+\left(1-{\hat{p}^{\mathcal{T}}}_k\right) \underbrace{\operatorname{KL}\left({\hat{\boldsymbol{p}}^{\mathcal{T}}}_{{\backslash k}} \| \hat{\boldsymbol{p}}^{\mathcal{S}}_{{\backslash k}}\right)}_{\mathrm{NCLD}},
\label{eq:kd1}
\end{aligned}
\end{equation}
where the TCLD term represents the similarity between $\boldsymbol{b}^{\mathcal{T}}$ and $\boldsymbol{b}^{\mathcal{S}}$ (binary probability vectors of teacher and student networks), while the NCLD term denotes the similarity between ${\hat{\boldsymbol{p}}^{\mathcal{T}}}_{{\backslash k}}$ and ${\hat{\boldsymbol{p}}^{\mathcal{S}}}_{{\backslash k}}$. Obviously, both TCLD and the weight of NCLD terms are related to $\hat{p}^{\mathcal{T}}_k$, making them coupled. As ($1-\hat{p}^{\mathcal{T}}_k$) is often much smaller than the weight of TCLD term (consistently 1), effects of NCLD term are often suppressed. However, in \cite{zhao2022decoupled}, authors claimed that the primary contribution of LD comes from NCLD term, and reformulated (\ref{eq:kd1}) as follows:
\begin{equation}
\mathrm{DKD}=\alpha \mathrm{TCLD}+\beta \mathrm{NCLD}.
\label{eq.dkd}
\end{equation}
Two independent hyper-parameters $\alpha$ and $\beta$ are used to balance the TCLD and NCLD terms \cite{zhao2022decoupled}. Extensive experiments conducted across a variety of datasets demonstrate that setting $\alpha$ to 1, as consistent with (\ref{eq:kd1}), and assigning $\beta$ positive integer values between 1 and 10 results in improved distillation performance compared to conventional LD (detailed in Sect. \ref{sec.exp_ld}). Nevertheless, upon revisiting (\ref{eq:kd1}), it becomes evident that ($1-\hat{p}^{\mathcal{T}}_k$) at each pixel varies independently. Thus, we propose a strategy to dynamically control the weight of the NCLD term across logits.

A logit value $z^{\mathcal{S}}_k$ approaching positive infinity indicates high confidence in the classification of $\boldsymbol{q}$. This implies that ambiguity arises in the student network when $z^{\mathcal{S}}_k$ is not sufficiently high. Differentiating DKD with respect to $z^{\mathcal{S}}_k$ yields the following expression: 
\begin{equation}
\begin{aligned}
\frac{\partial \mathrm{DKD}}{\partial z^{\mathcal{S}}_k } 
&=\alpha\frac{\partial \mathrm{TCLD}}{\partial z^{\mathcal{S}}_k}
+\beta\frac{\partial \mathrm{NCLD}}{\partial z^{\mathcal{S}}_k }\\
&=\alpha\frac{\mathrm{KL}\left(\boldsymbol{b}^{\mathcal{T}} \| \boldsymbol{b}^{\mathcal{S}}\right)}
{\partial z^{\mathcal{S}}_k}
+
\beta\frac{\operatorname{KL}\left({\hat{\boldsymbol{p}}^{\mathcal{T}}}_{{\backslash k}} \| \hat{\boldsymbol{p}}^{\mathcal{S}}_{{\backslash k}}\right)
}
{\partial z^{\mathcal{S}}_k }\\
&=\alpha \partial
\left( 
\hat{p}^{\mathcal{T}}_k 
\log \frac{\hat{p}^{\mathcal{T}}_k }{\hat{p}^{\mathcal{S}}_k }
+\left(1-\hat{p}^{\mathcal{T}}_k\right) 
\log \frac{1-\hat{p}^{\mathcal{T}}_k}{1-\hat{p}^{\mathcal{S}}_k}
\right)/{\partial z^{\mathcal{S}}_k}
\\
&=\alpha\left(-
\frac{\hat{p}^{\mathcal{T}}_k}{\hat{p}^{\mathcal{S}}_k}
\frac{\partial \hat{p}^{\mathcal{S}}_k}{\partial z^{\mathcal{S}}_k } 
+\frac{\left(1-\hat{p}^{\mathcal{T}}_k\right)}{\left(1-\hat{p}^{\mathcal{S}}_k\right)}\frac{\partial \hat{p}^{\mathcal{S}}_k}{\partial z^{\mathcal{S}}_k}
\right)+0,
\label{eq:derivation}
\end{aligned}
\end{equation}
which reveals that the gradient of the DKD loss with respect to $z^{\mathcal{S}}_k$ is only related to TCLD, and NCLD contributes zero gradients to the logit optimization. However, as discussed earlier, it is essential to pay more attention to NCLD, especially when the student network is less confident. Therefore, we are motivated to introduce the dynamic weight controller built upon the confidence of student's logits to dynamically balance TCLD and NCLD.

\subsubsection{\textbf{Dynamic Weight Controller}}
A thorough search of the relevant literature reveals no discussions on the adaptive control of the NCLD weight. We were inspired by a recent study \cite{guo2023boosting} that discussed setting adaptive temperature for KD in graph neural networks. However, this approach uniformly affects all logits, failing to account for the varying levels of confidence across different classes and instances. As depicted in Fig. \ref{fig.framework}, we assign each logit with an adaptive NCLD weight based on both the probability vector $\hat{\boldsymbol{p}}$ and the confidence $c=-\hat{\boldsymbol{p}}^\top \log\hat{\boldsymbol{p}}$ at the given pixel $\boldsymbol{q}$. Expanding to the entire logit tensor $\boldsymbol{Z}^{\mathcal{S}} \in \mathbb{R}^{C \times H \times W}$ generated by the student network, we first compute the confidence vector ${\boldsymbol{c}}^{\mathcal{S}}\in\mathbb{R}^{HW}$ for all pixels using the following expression:
\begin{equation}
\boldsymbol{c}^{\mathcal{S}}=-
\left(
\hat{\boldsymbol{P}}^{\mathcal{S}} 
\odot
\log \hat{\boldsymbol{P}}^{\mathcal{S}}
\right) 
\boldsymbol{1}_C, 
\end{equation}
where $\odot$ represents element-wise dot product between two  matrices, $\boldsymbol{1}_C$ is a $C$-entry column vector of ones, and $\hat{\boldsymbol{P}}^{\mathcal{S}} \in \mathbb{R}^{HW\times C}$ is a matrix calculated by applying the softmax function to $\boldsymbol{Z}^{\mathcal{S}}$ and reshaping it into a two-dimensional matrix.

Subsequently, we assign logits using an adaptive weight matrix $\boldsymbol{\beta}^{\mathcal{S}}\in\mathbb{R}^{HW\times C}$, which is constructed based on $\hat{\boldsymbol{P}}^{\mathcal{S}}$ and $\boldsymbol{c}^{\mathcal{S}}$. $\boldsymbol{\beta}^{\mathcal{S}}$ is subsequently obtained as follows:
\begin{equation}
\begin{aligned}
\boldsymbol{\beta}^{\mathcal{S}}=&(\beta_\mathrm{max}-\beta_\mathrm{min})
\operatorname{Sigmoid}\big(\operatorname{MLP}\big(\boldsymbol{\Omega}^{\mathcal{S}}\big)\big)\\
&+\beta_\mathrm{min}\boldsymbol{1}_{HW}{\boldsymbol{1}_C}^\top,
\label{eq.beta}
\end{aligned}
\end{equation}
where $\boldsymbol{\beta}^{\mathcal{S}}$ stores the NCLD weights with fixed range $[\beta_\mathrm{min}, \beta_\mathrm{max}]$. We empirically set $\beta_\mathrm{min}$ to 1 and $\beta_\mathrm{max}$ to 10, the selection of which has been discussed in DKD \cite{zhao2022decoupled}. The comparison experiments for the fixed $\boldsymbol{\beta}^{\mathcal{S}}$ in this interval are detailed in Sect. \ref{sec.exp_ld}. $\boldsymbol{\Omega}^{\mathcal{S}}$ in eq. \ref{eq.beta} is obtained using the following expression:
\begin{equation}
\boldsymbol{\Omega}^{\mathcal{S}}=\operatorname{Concat}\big(\hat{\boldsymbol{P}}^{\mathcal{S}}, \boldsymbol{c}^{\mathcal{S}}\big),
\label{eq.big_omega}
\end{equation}
and ``$\operatorname{Concat}$'' represents the operation to concatenate the matrix $\hat{\boldsymbol{P}}^{\mathcal{S}} \in \mathbb{R}^{HW\times C}$ and the confidence vector ${\boldsymbol{c}}^{\mathcal{S}}\in\mathbb{R}^{HW}$ to obtain $\boldsymbol{\Omega}^{\mathcal{S}} \in \mathbb{R}^{HW\times (C+1)}$.
Through extensive experiments detailed in Sect. \ref{sec.exp_ld}, we first validate the effectiveness of our proposed DWC, when $\boldsymbol{\Omega}^{\mathcal{S}}=\hat{{\boldsymbol{P}}}^{\mathcal{S}}$. Moreover, it achieves improved performance, when further incorporating $\boldsymbol{c}^{\mathcal{S}}$ into $\boldsymbol{\Omega}^{\mathcal{S}}$. Therefore, extending (\ref{eq.dkd}) to the entire image results in the logit distillation loss as follows:
\begin{equation}
\begin{aligned}
    \mathcal{L}_{{L}} 
        &=\alpha\underbrace{\boldsymbol{1}_{(HW)}^{\top}
        \bigg(
        \boldsymbol{B}^{\mathcal{S}} \odot\log 
        \left(
        \boldsymbol{B}^{\mathcal{T}}\oslash\boldsymbol{B}^{\mathcal{S}}
        \right)
        \bigg)
        \boldsymbol{1}_2}_{\text{TCLD}}
        \\
        &+\underbrace{\boldsymbol{1}_{(HW)}^{\top}
\bigg(
\boldsymbol{\beta}^{\mathcal{S}}
\odot
\hat{\boldsymbol{P}}_{\backslash k} ^{\mathcal{S}} \odot\log 
\left(
\hat{\boldsymbol{P}}_{\backslash k} ^{\mathcal{T}}
\oslash
\hat{\boldsymbol{P}}_{\backslash k} ^{\mathcal{S}}
\right)
\bigg)
\boldsymbol{1}_C}_{\text{logit-wise-weighted NCLD}},
\end{aligned}
\label{eq.lwwloss}
\end{equation}
where $\oslash$ denotes element-wise division. Let $\boldsymbol{T}^{\mathcal{S}} \in \mathbb{R}^{ HW \times C}$ and $\boldsymbol{N}^{\mathcal{S}}=\boldsymbol{1}_{HW} \boldsymbol{1}_{C}^\top- \boldsymbol{T}^{\mathcal{S}} \in \mathbb{R}^{ HW \times C}$ be the target and non-target ground-truth matrices in a boolean format, where each row is a binary vector corresponding to the given segmentation ground truth. Then, $\boldsymbol{B}^{\mathcal{T,S}} \in \mathbb{R}^{HW\times 2}$, the matrices that store the binary probabilities at each pixel, are obtained as follows:
\begin{equation}
\boldsymbol{B}^{\mathcal{T,S}}=\operatorname{Concat}\left(
{\hat{\boldsymbol{P}}^{\mathcal{T,S}}\odot\boldsymbol{T}^{\mathcal{T,S}}\boldsymbol{1}_C},
{\hat{\boldsymbol{P}}^{\mathcal{T,S}}\odot\boldsymbol{N}^{\mathcal{T,S}}\boldsymbol{1}_C}\right),
\end{equation}
where 
\begin{equation}
\hat{\boldsymbol{P}}_{\backslash k} ^{\mathcal{T,S}}=\frac
{\exp
\left(
\operatorname{Reshape}
\left(\boldsymbol{Z}^{\mathcal{T,S}}
\right)
-
\delta\boldsymbol{T}^{\mathcal{T,S}}
\right)}
{
\exp
\left(\operatorname{Reshape}
\left(\boldsymbol{Z}^{\mathcal{T,S}}\
\right)-\delta\boldsymbol{T}^{\mathcal{T,S}}
\right)
\boldsymbol{1}_{C}{\boldsymbol{1}_{C}}^\top}
\end{equation}
is a matrix of size $\mathbb{R}^{HW\times C}$ storing the independently modeled probabilities among non-target classes at each pixel, ``$\operatorname{Reshape}$'' operation denotes expanding a three-dimensional tensor into a two-dimensional matrix, and 
$\delta$ is a large value approaching infinity. Compared with (\ref{eq.dkd}), the inclusion of the variable $\boldsymbol{\beta}^{\mathcal{S}}$ in (\ref{eq.lwwloss}) allows each non-target class logit in the output to receive an adaptive weight that adjusts during the training process, thereby enhancing the overall performance of LD. The comprehensive quantitative comparison between DKD \cite{zhao2022decoupled} and our proposed DWLD is provided in Sect. \ref{sec.exp_ld}.

\subsection{Adaptively-Recalibrated Feature Distillation}
\label{sec:feature}
The above-mentioned logit distillation method relies exclusively on the output of the last layer, overlooking the significance of intermediate features within both teacher and student networks. It has nevertheless been demonstrated that these features play a critical role in effective representation learning, especially in deep neural networks \cite{romero2014FitNets}. Therefore, we design an adaptively-recalibrated feature distillation approach to further boost the transfer of spatial geometric prior knowledge from the teacher network to the student network. The remainder of this subsection details a feature recalibration process using kernel regression and a comprehensive feature consistency measurement method leveraging CKA \cite{nguyen2020wide} based on the HSIC \cite{ma2020hsic} (detailed in Fig. \ref{fig.framework}). 

\subsubsection{\textbf{Feature Recalibration via Kernel Regression}}
Let the feature maps produced by teacher and student networks be the tensors $\boldsymbol{F}^{\mathcal{T}}\in \mathbb{R}^{C^{\mathcal{T}} \times H^{\mathcal{T}} \times W^{\mathcal{T}}}$ and $\boldsymbol{F}^{\mathcal{S}}\in \mathbb{R}^{C^{\mathcal{S}} \times H^{\mathcal{S}} \times W^{\mathcal{S}}}$, respectively. Following the previous studies \cite{chen2021distilling,peng2019correlation,heo2019knowledge,yim2017gift}, we first align tensors $\boldsymbol{F}^{\mathcal{T}}$ and $\boldsymbol{F}^{\mathcal{S}}$ and produce matrix $\boldsymbol{F}_a^{\mathcal{T,S}}$ through the following process:
\begin{equation}
\boldsymbol{F}_a^{\mathcal{T,S}}=\operatorname{ReLU}\bigg(\operatorname{Norm}\Big(\underset{3 \times 3}{\operatorname { Conv }}\big(\operatorname{Reshape}(\boldsymbol{F}^{\mathcal{T,S}})\big)\Big)\bigg),
\label{eq.align}
\end{equation}
where $\boldsymbol{F}_a^{\mathcal{T,S}}\in \mathbb{R}^{C^{\mathcal{T}} \times HW}$, ``$\underset{3 \times 3}{\operatorname { Conv }}$'' denotes a convolutional layer built upon $3 \times 3$ kernels. and $HW=\min H^{\mathcal{T,S}}\min W^{\mathcal{T,S}}$. 
We take into account both spatial and channel alignment. As for features with mismatched resolutions, the ``$\operatorname{Reshape}$'' operation adjusts the larger feature map to match the smaller one.
Given the specific nature of our task, where feature maps from data-fusion and single-modal networks differ significantly in the channel dimension, the convolutional layer aligns the channels between the teacher and student features. Simultaneously, the activated and normalized features from the teacher and student networks share a more similar feature representation space.

Before delving into feature distillation, a pertinent question arises: are the features, having undergone alignment to a common shape using (\ref{eq.align}), ready for utilization? Extensive quantitative and qualitative experimental results lead us to a negative conclusion. We attribute the unsatisfactory performance of feature distillation to the discrepancy in feature scales and distributions between the teacher and student feature maps. On one hand, features often exhibit variations in different orders of magnitude, making their direct comparisons difficult \cite{tung2019similarity,park2019relational}. Working directly with the aligned features $\boldsymbol{F}_a^{\mathcal{T}}$ and $\boldsymbol{F}_a^{\mathcal{S}}$ could result in consistency measures being dominated by the model with large magnitude features. On the other hand, the absolute density of the data may vary significantly, and the shape of feature clusters may also differ depending on the locality \cite{aggarwal2015data}. Thus, recalibrating feature map distributions becomes another critical aspect that demands further attention. In particular, in our research problem, the teacher network learns heterogeneous features from both RGB images and spatial geometric information, while the student network is exclusively exposed to RGB images. Such a substantial difference in the learning data format exacerbates the knowledge transfer challenges. Taking inspiration from the knowledge transfer algorithm introduced in \cite{huang2017like}, which utilizes kernel regression to minimize the maximum mean discrepancy between teacher and student feature map probability distributions, we apply this technique to recalibrate the aligned features in our specific problem. The Laplace-based kernel regression process can be formulated as follows:
\begin{equation}
\boldsymbol{F}_k^{\mathcal{T,S}}=\operatorname{EXP} \left(-\frac{ (\boldsymbol{F}_a^{\mathcal{T,S}}- \bar{\boldsymbol{F}}_a^{\mathcal{T,S}}) \odot (\boldsymbol{F}_a^{\mathcal{T,S}}- \bar{\boldsymbol{F}}_a^{\mathcal{T,S}}) }{\sigma}\right),
\label{eq.lap}
\end{equation}
where ``$\operatorname{EXP}$'' represents the operation to take the exponent for each element in the tensor, and
\begin{equation}
\bar{\boldsymbol{F}}_a^{\mathcal{T,S}}=\frac{1}{H\times W\times C^{\mathcal{T}}}\boldsymbol{1}_{C^{\mathcal{T}}}^\top \boldsymbol{F}_a^{\mathcal{T,S}} \boldsymbol{1}_{HW}.
\end{equation}
The comprehensive comparisons of different kernel regression methods are given in Sect. \ref{sec.fd_ablation}.

\subsubsection{\textbf{Feature Consistency Measurement}}
As mentioned above, measuring the consistency (or similarity) between intermediate features in teacher and student networks is the key to feature distillation. Although Euclidean distance, cosine similarity, and the Pearson correlation coefficient can all serve this purpose, it has been witnessed that CKA \cite{nguyen2020wide} based on the HSIC \cite{ma2020hsic} provides a more comprehensive quantification of feature consistency, as demonstrated in several fundamental machine learning studies \cite{nguyen2020wide, kornblith2019similarity, cortes2012algorithms}.

We begin by computing two Gram matrices $\boldsymbol{T}_k={\boldsymbol{F}}_k^{\mathcal{T}} {{\boldsymbol{F}}_k^{\mathcal{T}}}^{\top}\in\mathbb{R}^{C^{\mathcal{T}}\times C^{\mathcal{T}}}$ and $\boldsymbol{S}_k={\boldsymbol{F}}_k^{\mathcal{S}} {{\boldsymbol{F}}_k^{\mathcal{S}}}^{\top}\in\mathbb{R}^{C^{\mathcal{T}}\times C^{\mathcal{T}}}$, reflecting the similarities between pairs of examples based on the representations contained in ${\boldsymbol{F}}_k^{\mathcal{T}}$ and ${\boldsymbol{F}}_k^{\mathcal{S}}$ \cite{nguyen2020wide}. Their HSIC measure is subsequently computed as follows \cite{ma2020hsic}:
\begin{equation}
\begin{aligned}
&\operatorname{HSIC}(\boldsymbol{T}_k,\boldsymbol{S}_k)  =\frac{{C^{\mathcal{T}}}^2}{(C^{\mathcal{T}}-1)^2}\bigg(\operatorname{tr}\left(\boldsymbol{S}_k \boldsymbol{T}_k\right)+
\\
&\frac{\boldsymbol{1}_{C^{\mathcal{T}}}^{\top} \boldsymbol{S} \boldsymbol{1}_{C^{\mathcal{T}}}  \boldsymbol {1}_{C^{\mathcal{T}}}^{\top} \boldsymbol{T}_k \boldsymbol{1}_{C^{\mathcal{T}}}}{{C^{\mathcal{T}}}^2}  -\frac{2}{{C^{\mathcal{T}}}^2} \boldsymbol{1}_{C^{\mathcal{T}}}^{\top} \boldsymbol{S}_k \boldsymbol{T}_k \boldsymbol{1}_{C^{\mathcal{T}}}\bigg),
\end{aligned}
\end{equation}
which statistically quantifies dependences between $\boldsymbol{T}_k$ and $\boldsymbol{S}_k$. CKA further normalizes the HSIC measures using the following expression:
\begin{equation}
\operatorname{CKA}(\boldsymbol{T}_k, \boldsymbol{S}_k)=\frac{\operatorname{HSIC}(\boldsymbol{T}_k, \boldsymbol{S}_k)}{\sqrt{\mathrm{HSIC}(\boldsymbol{T}_k, \boldsymbol{T}_k) \mathrm{HSIC}(\boldsymbol{S}_k, \boldsymbol{S}_k)}}.
\end{equation}
A CKA measure approaching 1 indicates that the intermediate features in teacher and student networks tend to be consistent. In our task, this suggests that the spatial geometric prior knowledge learned by the data-fusion teacher networks at the feature level is likely to have been implicitly infused into the single-modal student network. Therefore, the feature distillation loss can be formulated as follows: 
\begin{equation}
\begin{aligned}
\mathcal{L}_{F}=\sum_{n=1}^{N}(1-\operatorname{CKA}(\boldsymbol{T}_{k,n}, \boldsymbol{S}_{k,n})),
\label{eq.fd_loss}
\end{aligned}
\end{equation}
where $\boldsymbol{T}_{k,n}$ and $\boldsymbol{S}_{k,n}$ are yielded using the $n$-th pair of recalibrated teacher and student feature maps, respectively, and $N$ denotes the total number of feature maps. The quantitative comparisons of the above-mentioned feature consistency measurement methods are provided in Sect. \ref{sec.fd_ablation}. 

\subsection{Overall Loss}
\label{sec.loss}
As depicted in Fig. \ref{fig.framework}, our proposed LIX framework distills an RGB-X data-fusion teacher network into a student network trained exclusively with RGB images by two critical KD strategies: DWLD and ARFD. DWLD first reformulates the conventional KD loss into two independent and decoupled terms: TCLD and NCLD. Subsequently, a dynamic weight controller utilizing MLP layers generates a weight for each logit adaptively. Therefore, TCLD and weighted NCLD terms together form the LD loss $\mathcal{L}_L$. As for FD, we first utilize adaptive feature alignment and recalibration approaches based on kernel regression, which recalibrate the feature map distributions of teacher and student networks across various dimensions. Subsequently, we measure the similarity between features to formulate our novel FD loss $\mathcal{L}_F$ in teacher and student networks based on the CKA algorithm, which quantifies the feature consistency between teacher and student networks. 

In summary, the overall loss function can be formulated as a combination of the initial loss $\mathcal{L}_H$ of the hard labels (ground truth) and distillation losses, consisting of a logit distillation loss $\mathcal{L}_L$ and a feature distillation loss $\mathcal{L}_F$, as follows:
\begin{equation}
\mathcal{L}=
\mathcal{L}_H+
\lambda_L \mathcal{L}_L+
\lambda_F \mathcal{L}_F,
\label{eq.loss}
\end{equation}
where $\lambda_L$ and $\lambda_F$ are hyper-parameters to balance distillation losses. By minimizing (\ref{eq.loss}), the single-model student network can be implicitly infused with the spatial geometric knowledge learned by the data-fusion teacher network.

\section{Experiments}
\label{sec:experiments}
\subsection{Experimental Setup}
\begin{figure*}[t!]
		\centering
		\includegraphics[width=0.99\textwidth]{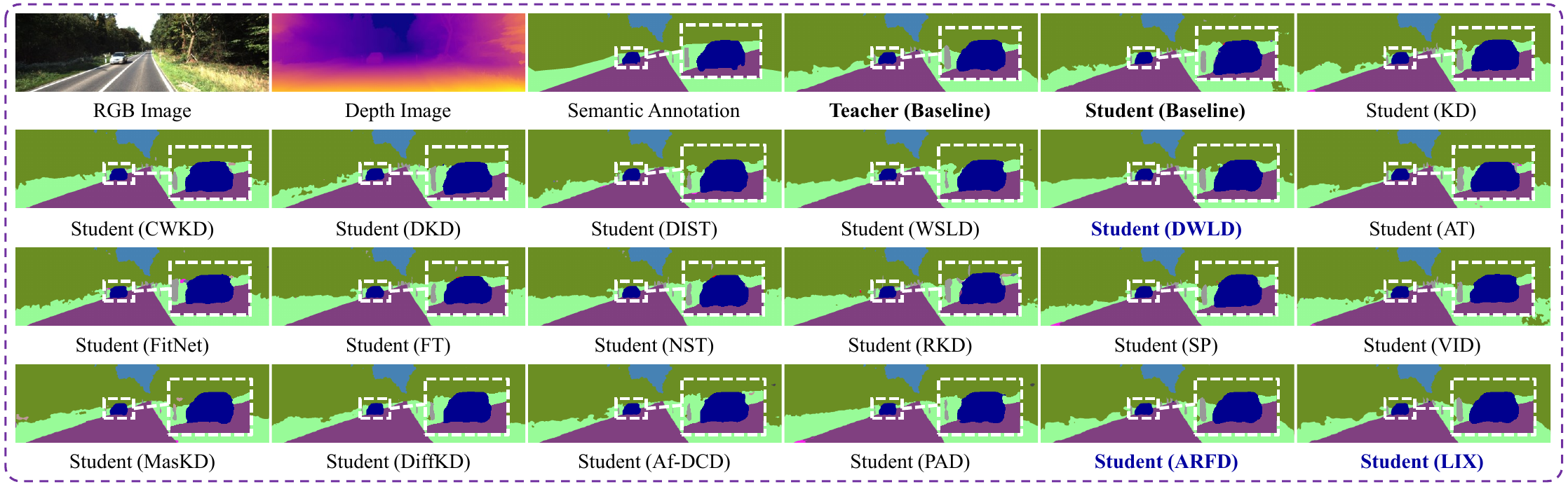}
		\centering
		\caption{Qualitative comparison with other SoTA KD approaches on the KITTI Semantics dataset \cite{menze2015kitti} using SNR-RoadSeg, where significantly improved areas are highlighted with white dashed boxes.
            }
		\label{fig.kitts}
\end{figure*}

\begin{figure*}[h!]
		\centering
		\includegraphics[width=0.99\textwidth]{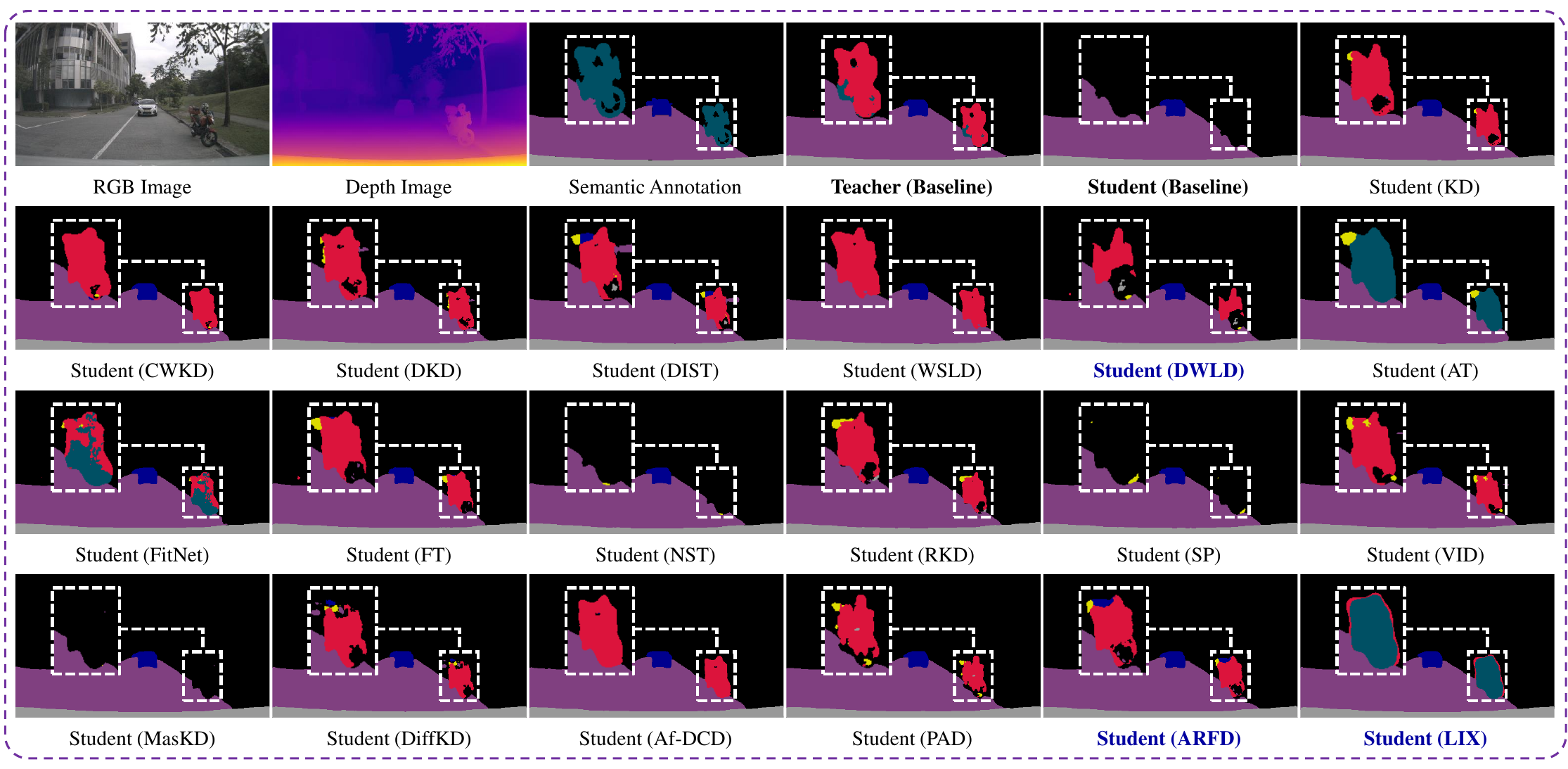}
		\centering
		\caption{Qualitative comparison with other SoTA KD approaches on the nuImage dataset \cite{nuscenes2019} using SNR-RoadSeg, where significantly improved areas are highlighted with white dashed boxes.
            }
		\label{fig.nuimg}
\end{figure*}

\subsubsection{\textbf{Datasets}}
We utilize three public datasets to evaluate the performance of our proposed methods: the vKITTI2 dataset \cite{cabon2020vkitti2} (synthetic yet large-scale), the KITTI Semantics dataset \cite{menze2015kitti} (real-world yet modest-scale) and the nuImage dataset \cite{nuscenes2019} (real-world yet large-scale). Their details are as follows:
\begin{itemize}
\item 
The \uline{\textbf{vKITTI2 dataset}} contains virtual replicas of five sequences from the KITTI dataset and provides semantic annotations for 15 different classes. Dense ground-truth depth maps are acquired through depth rendering using a virtual engine. In our experiments, we randomly select 700 images from this dataset, along with their semantic and depth annotations, to validate the effectiveness of our proposed logit and feature distillation approaches. These images are randomly divided into a training set and a validation set with a ratio of 5:2.

\item The \uline{\textbf{KITTI Semantics dataset}} contains 200 real-world images captured in various driving scenarios. It provides ground-truth semantic annotations for 19 different classes (in alignment with the Cityscapes \cite{cordts2016cityscapes} dataset). Sparse disparity ground truth is obtained using a Velodyne HDL-64E LiDAR. We generate dense depth maps using a well-trained CreStereo \cite{li2022practical} network in the experiments. These images are randomly divided into a training set and a validation set with a ratio of 3:1.

\item The \uline{\textbf{nuImage dataset}} is a large-scale, real-world dataset designed for autonomous driving perception. It consists of 93,476 images with 2M labeled objects. In our experiments, we randomly select 800 images with semantic annotations and generate dense depth maps using a pre-trained Depth Anything network \cite{yang2024depth}. These images are then randomly split into a training set and a validation set with a ratio of 7: 3.
\end{itemize}

\begin{table*}[!t]
\begin{center}
\settablefont
\caption{Comparison with SoTA knowledge distillation approaches on the vKITTI2 dataset.}
\label{tb.vk}
\begin{tabular}{c|c|cccc|cccc}
\toprule
\multirow{2}*{KD Type}
& \multirow{2}*{Algorithm}
&\multicolumn{4}{c|}{SNE-RoadSeg}
&\multicolumn{4}{c}{MFNet}\\
\cline{3-10}
& &mFsc (\%) $\uparrow$ &fwFsc (\%) $\uparrow$ &mIoU (\%) $\uparrow$ &fwIoU (\%) $\uparrow$
&mFsc (\%) $\uparrow$ &fwFsc (\%) $\uparrow$ &mIoU (\%) $\uparrow$ &fwIoU (\%) $\uparrow$\\
\hline
\hline
\multirow{2}*{Baseline}
&\cellcolor{gray!15}Teacher Network&\cellcolor{gray!15}97.72&\cellcolor{gray!15}99.12&\cellcolor{gray!15}95.63&\cellcolor{gray!15}98.27&\cellcolor{gray!15}{92.65}&\cellcolor{gray!15}{97.68}&\cellcolor{gray!15}{86.96}&\cellcolor{gray!15}{95.57}\\
&\cellcolor{gray!15}Student Network&\cellcolor{gray!15}94.51&\cellcolor{gray!15}97.83&\cellcolor{gray!15}89.91&\cellcolor{gray!15}95.82&\cellcolor{gray!15}91.22&\cellcolor{gray!15}96.93&\cellcolor{gray!15}84.59&\cellcolor{gray!15}94.19\\
\hline
\multirow{6}*{LD}
&KD\cite{hinton2015distilling}&94.89&97.90&90.59&95.94&91.93&97.06&85.70&94.42\\
&CWKD\cite{shu2021channel}&95.02&97.95&90.82&96.03&91.63&97.02&85.29&94.36\\
&DKD\cite{zhao2022decoupled}&94.97&97.94&90.75&95.97&91.68&97.06&85.29&94.42\\
&DIST\cite{huang2022knowledge}&95.17&97.97&91.08&96.06&84.21&95.05&74.58&91.01\\
&\clr{WSLD\cite{zhourethinking}}&\clr{94.86}&\clr{97.95}&\clr{90.52}&\clr{96.04}&\clr{85.85}&\clr{95.38}&\clr{76.69}&\clr{91.53}\\
&\cellcolor{green!15}\textbf{DWLD(Ours)}&\textbf{95.42}\cellcolor{green!15}&\textbf{98.06}\cellcolor{green!15}&\textbf{91.50}\cellcolor{green!15}&\textbf{96.24}\cellcolor{green!15}&\textbf{92.29}\cellcolor{green!15}&\textbf{97.22}\cellcolor{green!15}&\textbf{85.94}\cellcolor{green!15}&\textbf{94.68}\cellcolor{green!15}\\
\hline
\multirow{12}*{FD}
&AT\cite{zagoruyko2016paying}&94.61&97.87&90.12&95.89&91.97&97.13&85.78&94.55\\
&FitNet\cite{romero2014FitNets}&94.89&97.92&90.59&95.99&91.36&96.93&84.83&94.19\\
&FT\cite{kim2018paraphrasing}&94.75&97.91&90.34&95.97&91.47&96.96&84.96&94.25\\
&NST\cite{huang2017like}&83.19&90.94&73.27&84.55&89.39&96.58&81.88&93.58\\
&RKD\cite{park2019relational}&94.69&97.81&90.25&95.78&85.33&95.44&76.19&91.66\\
&SP\cite{tung2019similarity}&95.08&97.96&90.93&96.06&91.98&97.10&85.78&94.50\\
&VID\cite{ahn2019variational}&95.05&97.95&90.87&96.03&91.78&97.01&85.54&94.34\\
&\clr{MasKD\cite{huangmasked}}&\clr{81.26}&\clr{92.07}&\clr{70.95}&\clr{86.34}&\clr{80.92}&\clr{94.08}&\clr{70.45}&\clr{89.44}\\
&\clr{DiffKD\cite{huang2023knowledge}}&\clr{94.87}&\clr{97.92}&\clr{90.54}&\clr{95.98}&\clr{84.67}&\clr{94.93}&\clr{75.17}&\clr{90.75}\\
&\clr{Af-DCD\cite{fan2023augmentation}}&\clr{93.81}&\clr{97.72}&\clr{88.76}&\clr{95.62}&\clr{80.34}&\clr{94.21}&\clr{70.00}&\clr{89.76}\\
&\clr{PAD\cite{zhang2020prime}}&\clr{95.11}&\clr{97.97}&\clr{90.96}&\clr{96.07}&\clr{85.63}&\clr{95.07}&\clr{76.30}&\clr{90.96}\\
&\cellcolor{blue!15}\textbf{ARFD(Ours)}&\cellcolor{blue!15}\textbf{95.20}&\textbf{98.01}\cellcolor{blue!15}&\textbf{90.97}\cellcolor{blue!15}&\textbf{96.16}\cellcolor{blue!15}&\textbf{92.46}\cellcolor{blue!15}&\textbf{97.27}\cellcolor{blue!15}&\textbf{86.52}\cellcolor{blue!15}&\textbf{94.80}\cellcolor{blue!15}\\
\hline
LD+FD&\cellcolor{red!15}\textbf{LIX(Ours)}&\textbf{95.79}\cellcolor{red!15}&\textbf{98.23}\cellcolor{red!15}&\textbf{91.95}\cellcolor{red!15}&\textbf{96.33}\cellcolor{red!15}&\textbf{92.50}\cellcolor{red!15}&\textbf{97.32}\cellcolor{red!15}&\textbf{86.88}\cellcolor{red!15}&\textbf{95.10}\cellcolor{red!15}\\
\bottomrule
\end{tabular}
\end{center}
\end{table*}

\begin{table*}[!t]
\begin{center}
\settablefont
\caption{Comparison with SoTA knowledge distillation approaches on the KITTI Semantics dataset.}
\label{tb.ks}
\begin{tabular}{c|c|cccc|cccc}
\toprule
\multirow{2}*{KD Type}
& \multirow{2}*{Algorithm}
&\multicolumn{4}{c|}{SNE-RoadSeg}
&\multicolumn{4}{c}{MFNet}\\
\cline{3-10}
& &mFsc (\%) $\uparrow$ &fwFsc (\%) $\uparrow$ &mIoU (\%) $\uparrow$ &fwIoU (\%) $\uparrow$
&mFsc (\%) $\uparrow$ &fwFsc (\%) $\uparrow$ &mIoU (\%) $\uparrow$ &fwIoU (\%) $\uparrow$\\
\hline
\hline
\multirow{2}*{Baseline}
& \cellcolor{gray!15} Teacher Network & \cellcolor{gray!15}70.85&\cellcolor{gray!15}93.02&\cellcolor{gray!15}61.34&\cellcolor{gray!15}87.76&45.17\cellcolor{gray!15}&89.22\cellcolor{gray!15}&37.86\cellcolor{gray!15}&82.37\cellcolor{gray!15}\\
&\cellcolor{gray!15}Student Network&\cellcolor{gray!15}59.43&\cellcolor{gray!15}90.48&\cellcolor{gray!15}49.17&\cellcolor{gray!15}83.86&35.91\cellcolor{gray!15}&86.52\cellcolor{gray!15}&30.68\cellcolor{gray!15}&79.19\cellcolor{gray!15}\\
\hline
\multirow{6}*{LD}
&KD\cite{hinton2015distilling}&\textbf{63.71}&90.88&52.75&84.39&37.38&87.14&31.79&80.01\\
&CWKD\cite{shu2021channel}&60.84&90.49&50.15&83.88&38.04&87.14&32.25&79.88\\
&DKD\cite{zhao2022decoupled}&61.61&90.57&51.28&83.90&37.62&86.35&31.61&78.65\\
&\clr{DIST\cite{huang2022knowledge}}&\clr{55.26}&\clr{91.71}&\clr{47.71}&\clr{85.88}&37.94&86.74&32.05&79.31\\
&\clr{WSLD\cite{zhourethinking}}&\clr{53.24}&\clr{91.16}&\clr{45.66}&\clr{85.09}&\textbf{40.57}&87.77&33.91&80.49\\
&\cellcolor{green!15}\textbf{DWLD(Ours)}&62.78\cellcolor{green!15}&\textbf{92.36}\cellcolor{green!15}&\textbf{53.54}\cellcolor{green!15}&\textbf{86.62}\cellcolor{green!15}&39.82\cellcolor{green!15}&\textbf{88.41}\cellcolor{green!15}&\textbf{33.93}\cellcolor{green!15}&\textbf{81.62}\cellcolor{green!15}\\
\hline
\multirow{12}*{FD}
&AT\cite{zagoruyko2016paying}&61.81&91.24&51.57&84.98&37.02&86.36&31.31&78.80\\
&FitNet\cite{romero2014FitNets}&57.88&90.46&48.18&83.82&38.70&87.29&32.88&80.02\\
&FT\cite{kim2018paraphrasing}&60.61&90.41&50.16&83.71&38.24&86.89&32.37&79.52\\
&NST\cite{huang2017like}&55.41&87.77&45.56&79.93&40.53&\textbf{88.08}&34.05&\textbf{81.06}\\
&RKD\cite{park2019relational}&58.52&89.55&47.74&82.50&33.11&84.26&28.10&76.21\\
&SP\cite{tung2019similarity}&\textbf{62.89}&90.52&52.11&83.91&38.35&86.92&32.49&79.51\\
&VID\cite{ahn2019variational}&62.45&90.51&51.66&83.89&38.16&87.07&32.32&79.78\\
&\clr{MasKD\cite{huangmasked}}&\clr{51.41}&\clr{88.58}&\clr{42.32}&\clr{81.56}&40.85&87.27&34.10&79.90\\
&\clr{DiffKD\cite{huang2023knowledge}}&\clr{55.94}&\clr{91.97}&\clr{48.37}&\clr{86.29}&\textbf{41.48}&87.73&34.45&80.29\\
&\clr{Af-DCD\cite{fan2023augmentation}}&\clr{51.04}&\clr{91.65}&\clr{43.60}&\clr{85.81}&41.02&86.75&33.74&78.82\\
&\clr{PAD\cite{zhang2020prime}}&\clr{48.65}&\clr{90.84}&\clr{41.09}&\clr{84.65}&36.99&86.70&31.65&79.40\\
&\cellcolor{blue!15}\textbf{ARFD(Ours)}&62.25\cellcolor{blue!15}&\textbf{92.34}\cellcolor{blue!15}&\textbf{52.92}\cellcolor{blue!15}&\textbf{86.59}\cellcolor{blue!15}&41.32\cellcolor{blue!15}&87.76\cellcolor{blue!15}&\textbf{34.46}\cellcolor{blue!15}&80.56\cellcolor{blue!15}\\
\hline
LD+FD&\cellcolor{red!15}\textbf{LIX(Ours)}&\textbf{63.70}\cellcolor{red!15}&\textbf{92.47}\cellcolor{red!15}&\textbf{54.65}\cellcolor{red!15}&\textbf{86.80}\cellcolor{red!15}&\textbf{43.25}\cellcolor{red!15}&\textbf{88.59}\cellcolor{red!15}&\textbf{36.30}\cellcolor{red!15}&\textbf{81.63}\cellcolor{red!15}\\
\bottomrule
\end{tabular}
\end{center}
\end{table*}
\subsubsection{\textbf{Implementation Details and Evaluation Metrics}}
Our experiments are conducted on an NVIDIA RTX 3090 GPU. \clr{All images in the vKITTI2 dataset and the KITTI Semantics dataset are resized to $1,248\times384$ pixels before being fed into the network, and images in the nuImage dataset are resized to $512\times288$ pixels.} We utilize the stochastic gradient descent (SGD) \cite{robbins1951stochastic} optimizer for network training, with momentum and weight decay parameters set to 0.9 and $5\times10^{-4}$, respectively. The initial learning rate is set to $5\times 10^{-3}$, and training is conducted for 500 epochs with early stopping used to prevent over-fitting. \clr{To ensure stable network training, the batch size is set to 3 for SNE-RoadSeg on the vKITTI2 dataset and KITTI Semantics dataset, 8 for MFNet on the vKITTI dataset and KITTI Semantics, and 8 for both networks on the nuImage dataset. To enhance the model's robustness, we have employed standard data augmentation techniques, such as random color adjustment, photometric distortion, rescaling, and flipping. We employ four metrics to quantify the performance of the KD algorithms: the mean and frequency-weighted F1-score (abbreviated as mFsc and fwFsc, respectively) as well as the mean and frequency-weighted intersection over union (abbreviated as mIoU and fwIoU, respectively).}

\subsection{Comparison with State-of-the-Art Methods}
\label{sec.compare_with_sota}

\begin{table*}[!t]
\begin{center}
\settablefont
\caption{Comparison with SoTA knowledge distillation approaches on the nuImage dataset.}
\label{tb.nuimg}
\begin{tabular}{c|c|cccc|cccc}
\toprule
\multirow{2}*{KD Type}
& \multirow{2}*{Algorithm}
&\multicolumn{4}{c|}{SNE-RoadSeg}
&\multicolumn{4}{c}{MFNet}\\
\cline{3-10}
& &mFsc (\%) $\uparrow$ &fwFsc (\%) $\uparrow$ &mIoU (\%) $\uparrow$ &fwIoU (\%) $\uparrow$
&mFsc (\%) $\uparrow$ &fwFsc (\%) $\uparrow$ &mIoU (\%) $\uparrow$ &fwIoU (\%) $\uparrow$\\
\hline
\hline
\multirow{2}*{Baseline}
&\cellcolor{gray!15}Teacher Network&\cellcolor{gray!15}74.42&\cellcolor{gray!15}96.83&\cellcolor{gray!15}65.80&\cellcolor{gray!15}94.16&59.98\cellcolor{gray!15}&94.59\cellcolor{gray!15}&54.06\cellcolor{gray!15}&90.40\cellcolor{gray!15}\\
&\cellcolor{gray!15}Student Network&\cellcolor{gray!15}60.83&\cellcolor{gray!15}95.30&\cellcolor{gray!15}55.27&\cellcolor{gray!15}91.66&55.97\cellcolor{gray!15}&92.51\cellcolor{gray!15}&49.03\cellcolor{gray!15}&87.03\cellcolor{gray!15}\\
\hline
\multirow{6}*{LD}
&KD\cite{hinton2015distilling}&68.20&95.61&60.46&92.07&56.11&92.98&49.35&87.79\\
&CWKD\cite{shu2021channel}&67.17&95.62&59.48&92.11&56.49&93.01&49.72&87.84\\
&DKD\cite{zhao2022decoupled}&67.66&95.60&60.06&92.04&56.38&92.92&49.59&87.72\\
&\clr{DIST\cite{huang2022knowledge}}&67.11&95.56&59.50&91.99&56.38&92.97&49.61&87.77\\
&\clr{WSLD\cite{zhourethinking}}&67.16&95.53&59.46&91.96&57.52&93.18&50.00&88.12\\
&\cellcolor{green!15}\textbf{DWLD(Ours)}&\textbf{68.70}\cellcolor{green!15}&\textbf{95.83}\cellcolor{green!15}&\textbf{61.11}\cellcolor{green!15}&\textbf{92.45}\cellcolor{green!15}&\textbf{57.55}\cellcolor{green!15}&\textbf{93.21}\cellcolor{green!15}&\textbf{50.67}\cellcolor{green!15}&\textbf{88.15}\cellcolor{green!15}\\
\hline
\multirow{12}*{FD}
&AT\cite{zagoruyko2016paying}&66.06&95.42&58.69&91.79&57.02&93.16&50.15&\textbf{88.09}\\
&FitNet\cite{romero2014FitNets}&\textbf{70.54}&95.66&61.39&92.11&56.28&92.92&49.46&87.71\\
&FT\cite{kim2018paraphrasing}&67.21&95.76&59.52&92.36&56.33&92.99&49.56&87.80\\
&NST\cite{huang2017like}&62.24&\textbf{95.88}&56.21&\textbf{92.68}&\textbf{57.39}&\textbf{93.26}&50.06&87.26\\
&RKD\cite{park2019relational}&64.87&95.02&57.07&91.11&53.37&91.59&46.45&85.62\\
&SP\cite{tung2019similarity}&61.45&95.41&55.90&91.83&56.27&92.93&49.45&87.70\\
&VID\cite{ahn2019variational}&68.29&95.65&60.64&92.13&56.38&92.90&49.45&87.65\\
&\clr{MasKD\cite{huangmasked}}&62.48&95.68&57.08&92.32&57.39&92.26&50.05&87.22\\
&\clr{DiffKD\cite{huang2023knowledge}}&66.73&95.32&58.92&91.59&56.78&93.02&50.07&87.85\\
&\clr{Af-DCD\cite{fan2023augmentation}}&66.14&94.99&58.18&91.01&57.53&93.31&50.08&87.34\\
&\clr{PAD\cite{zhang2020prime}}&66.42&95.21&58.52&91.38&56.87&92.85&49.94&87.57\\
&\cellcolor{blue!15}\textbf{ARFD(Ours)}&69.49\cellcolor{blue!15}&\textbf{95.88}\cellcolor{blue!15}&\textbf{61.84}\cellcolor{blue!15}&92.52\cellcolor{blue!15}&56.94\cellcolor{blue!15}&93.04\cellcolor{blue!15}&\textbf{50.26}\cellcolor{blue!15}&87.89\cellcolor{blue!15}\\
\hline
LD+FD&\cellcolor{red!15}\textbf{LIX(Ours)}&\textbf{72.90}\cellcolor{red!15}&\textbf{96.06}\cellcolor{red!15}&\textbf{63.54}\cellcolor{red!15}&\textbf{92.79}\cellcolor{red!15}&\textbf{57.91}\cellcolor{red!15}&\textbf{93.34}\cellcolor{red!15}&\textbf{51.06}\cellcolor{red!15}&\cellcolor{red!15}\textbf{88.38}\\
\bottomrule
\end{tabular}
\end{center}
\end{table*}

\clr{The quantitative and qualitative experimental results on the vKITTI2, KITTI Semantics, and nuImage datasets are presented in Tables \ref{tb.vk}, \ref{tb.ks} and \ref{tb.nuimg} as well as Figs. \ref{fig.kitts} and \ref{fig.nuimg}. In our baseline experiments, the teacher network utilizes a duplex encoder for RGB-D semantic segmentation, while the student network uses a single encoder to learn semantic clues exclusively from RGB images. These results suggest that the student network of our proposed LIX achieves over 90\% of the mFsc, and in some cases, is nearly comparable to that of the teacher network. Moreover, when DWLD is utilized solely in this specific task, we observe significant improvements compared to the baseline student network.} These results serve as strong evidence for the effectiveness and superior performance of our proposed DWLD algorithm. We arrive at a similar conclusion when evaluating the performance of ARFD. LIX, the combined use of these two algorithms enables the single-modal student network to achieve comparable performance to that of the data-fusion teacher network. This demonstrates the effectiveness of our proposed KD strategy for the implicit infusion of spatial geometric prior knowledge. 

Moreover, we observe that KD's performance is significantly influenced by the network parameters as well as the difficulty level of the dataset. The encoders of the teacher and student SNE-RoadSeg networks have 116.28 M and 58.14 M parameters, respectively, while the encoders of the teacher and student MFNet networks have 0.60 M and 0.52 M parameters, respectively. \clr{As illustrated in Table \ref{tb.vk}, since the vKITTI2 dataset is larger and less challenging, DWLD, ARFD, and LIX all achieve the SoTA performance regardless of the network parameters.} When our method is applied to MFNet, its performance is on par with that of the teacher network on the vKITTI2 dataset, demonstrating superior effectiveness compared to SNE-RoadSeg with LIX. We attribute this superior performance not only to the comparable parameter numbers between the teacher and student MFNet networks but also to the lower difficulty level of the dataset. It is possible that the lightweight MFNet is adequately capable of learning semantic segmentation on this less challenging dataset.

On the other hand, the experimental results on the KITTI Semantics dataset have exceeded our expectations, when using MFNet in conjunction with our method. \clr{As illustrated in Table \ref{tb.ks}, ARFD outperforms other algorithms only in mIoU and achieves performance comparable to that of the SoTA algorithms when evaluated using other metrics. We attribute this to a potential ``mismatch'' between MFNet and the KITTI Semantics dataset, where a lightweight network may struggle to handle a challenging dataset. Additionally, while the student SNE-RoadSeg network trained via LIX achieves superior performance, its mIoU reaches only 89\% of the teacher network's mIoU. We believe that infusing spatial geometric prior knowledge into the student network is effective but it fails in some cases. For instance, if teacher and student networks have significantly different numbers of model parameters, the student network may struggle with challenging datasets.}

\begin{figure*}[h!]
		\centering
		\includegraphics[width=0.98\textwidth]{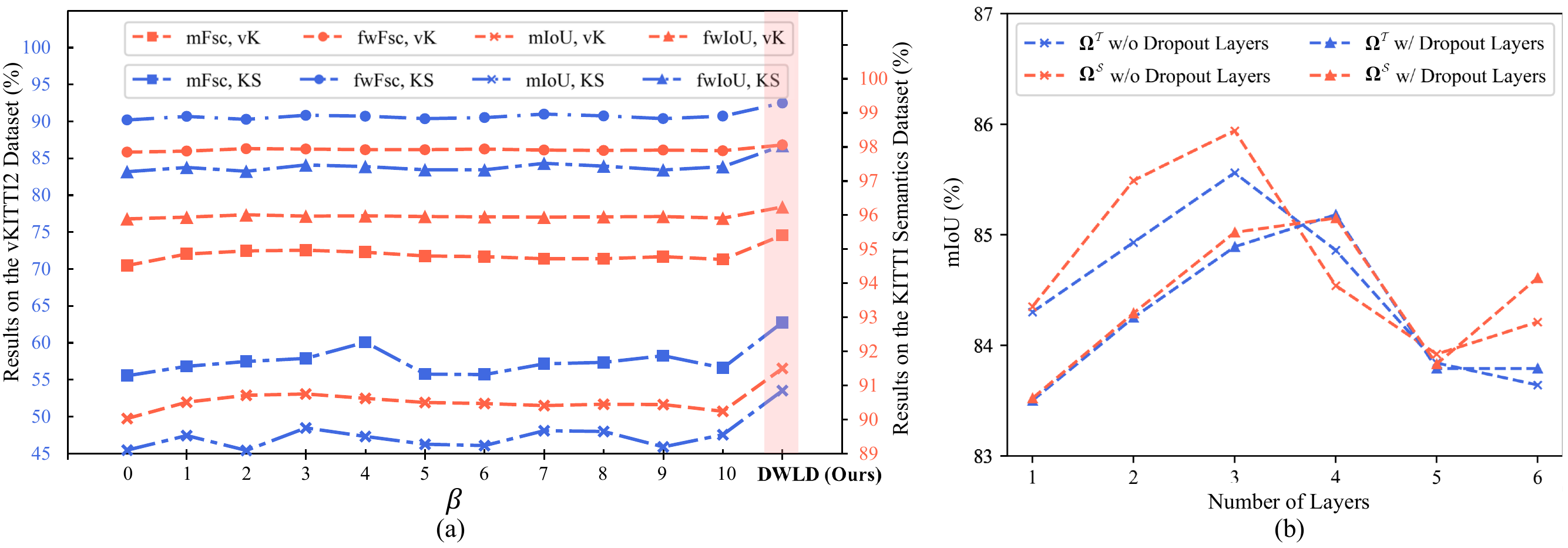}
		\centering
		\caption{\clr{Ablation studies on LD: (a) comparison between DKD \cite{zhao2022decoupled} and DWLD with respect to different $\beta$, where ``vK'' and ``KS'' are the abbreviations of ``vKITTI2'' and ``KITTI Semantics'', respectively; (b) comparison among various designs of $\boldsymbol{\Omega}^{\mathcal{S}}$ on the vKITTI2 dataset.}}
		\label{fig.ld}
\end{figure*} 

\clr{
As depicted in Table \ref{tb.nuimg}, the experimental results on the nuImage dataset highlight LIX's adaptability to real-world scenarios with rich contextual diversity. Notably, LIX demonstrates superior performance with a 12\% increase in terms of mFsc over the baseline student network. This success can be attributed to the large-scale and diverse nature of the nuImage dataset, which provides sufficient samples for the student SNE-RoadSeg network to effectively absorb the teacher's spatial geometric prior knowledge. Nevertheless, as shown in Fig. \ref{fig.nuimg}, only student networks trained via LIX and AT \cite{zagoruyko2016paying} can recognize the motorcycle, while the remaining networks misclassify it as a person. This failure may stem from the teacher network misleads the student network during the distillation process. Notably, LIX successfully achieves accurate segmentation results, demonstrating the collaborative effect of combining logit and feature distillation techniques in a challenging large-scale dataset. Additionally, the improvements achieved by KD methods using MFNet are relatively modest. With fewer parameters to handle extensive data, the teacher's knowledge is inherently limited, leaving less room for the student to improve. 
}

\clr{
While existing KD methods can be effectively applied to solve this problem, their performance generally falls short of our newly proposed LIX framework, which is specifically developed for this task. Among LD methods, classical KD \cite{hinton2015distilling} and CWKD \cite{shu2021channel} demonstrate moderate improvements over the student baseline but lag behind more advanced approaches \cite{zhao2022decoupled}. Furthermore, our proposed DWLD focuses on uncertain logits by assigning an appropriate weight to each logit, leading to a 0.02-9.25\% higher IoU compared to WSLD, which prioritizes soft label regularization. As for FD methods, AT \cite{zagoruyko2016paying} and FitNet \cite{romero2014FitNets} yield only marginal improvements, highlighting the challenge of direct feature alignment in complex scenarios. Methods such as SP \cite{tung2019similarity} and VID \cite{ahn2019variational} achieve better results by preserving pairwise feature similarities. However, their reliance on dimensionality reduction limits their effectiveness in fine-grained scene parsing. 
}

\subsection{Ablation Study on Logit Distillation}
\label{sec.exp_ld}

\begin{table}[!t]
\begin{center}
\settablefont
\caption{Comparison among various designs of $\boldsymbol{\Omega}^{\mathcal{S}}$ on the vKITTI2 dataset.}
\label{tb.omega}
\begin{tabular}{c|cccc}
\toprule
$\boldsymbol{\Omega}^{\mathcal{S}}$ Type &mFsc (\%)    &fwFsc (\%)    &mIoU (\%)    &fwIoU (\%)  \\
\hline
\hline
$\hat{\boldsymbol{P}}^{\mathcal{S}}$ & 90.90&96.84 &84.12 &94.03 \\
$\boldsymbol{c}^{\mathcal{S}}$ &91.42 &96.93 &84.96 &94.22 \\
$\operatorname{Concat}\big(\hat{\boldsymbol{P}}^{\mathcal{S}}, \boldsymbol{c}^{\mathcal{S}}\big)$ &\textbf{92.29} &\textbf{97.22} &\textbf{85.94} &\textbf{94.68} \\
\bottomrule
\end{tabular}
\end{center}
\end{table}

\begin{figure}[!t]
		\centering
		\includegraphics[width=0.49\textwidth]{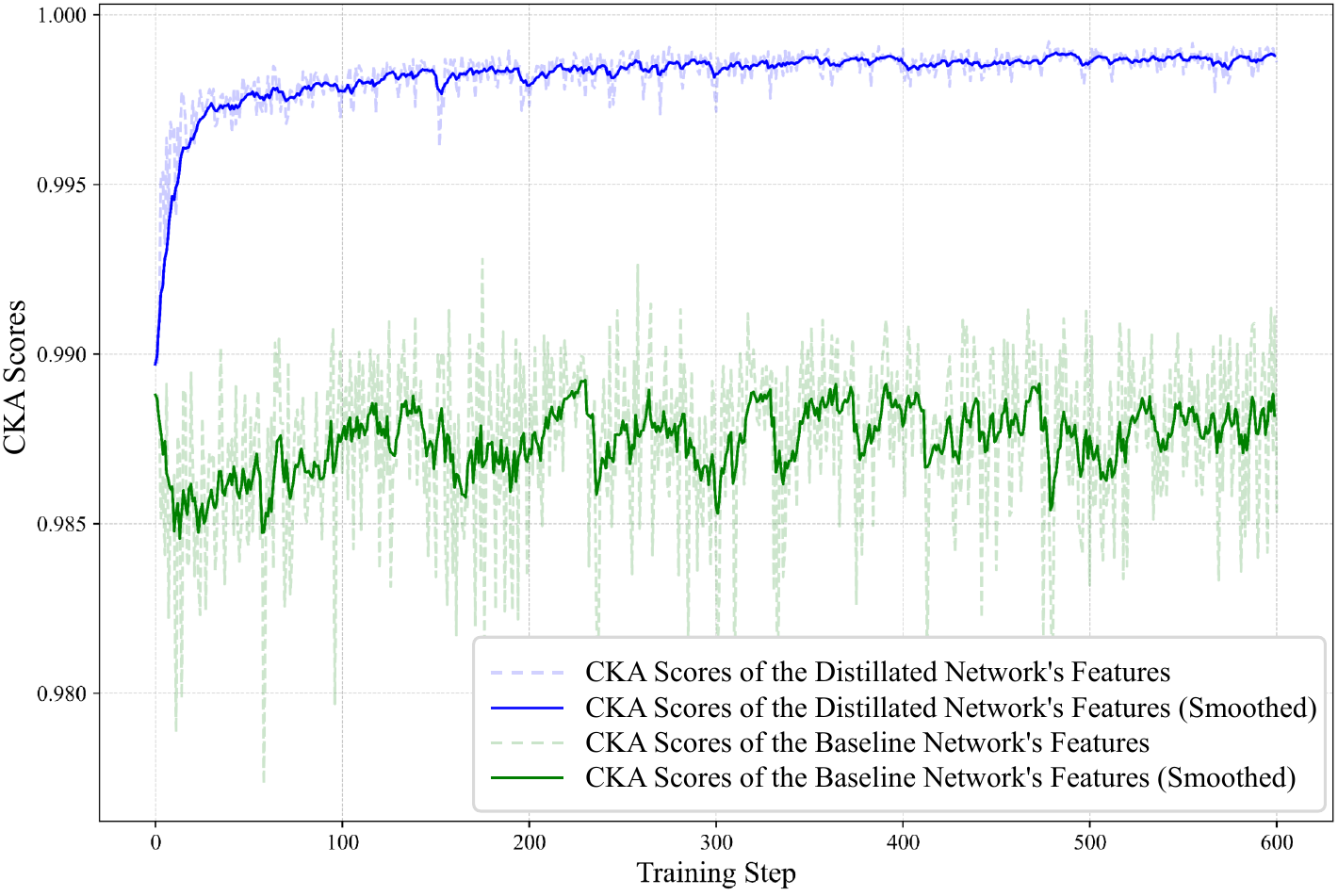}
		\centering
		\caption{\clr{Comparison of the CKA scores between the features of baseline and distilled student networks.}         
            }
		\label{fig.cka}
\end{figure}

\begin{figure*}[h!]
		\centering
		\includegraphics[width=0.99\textwidth]{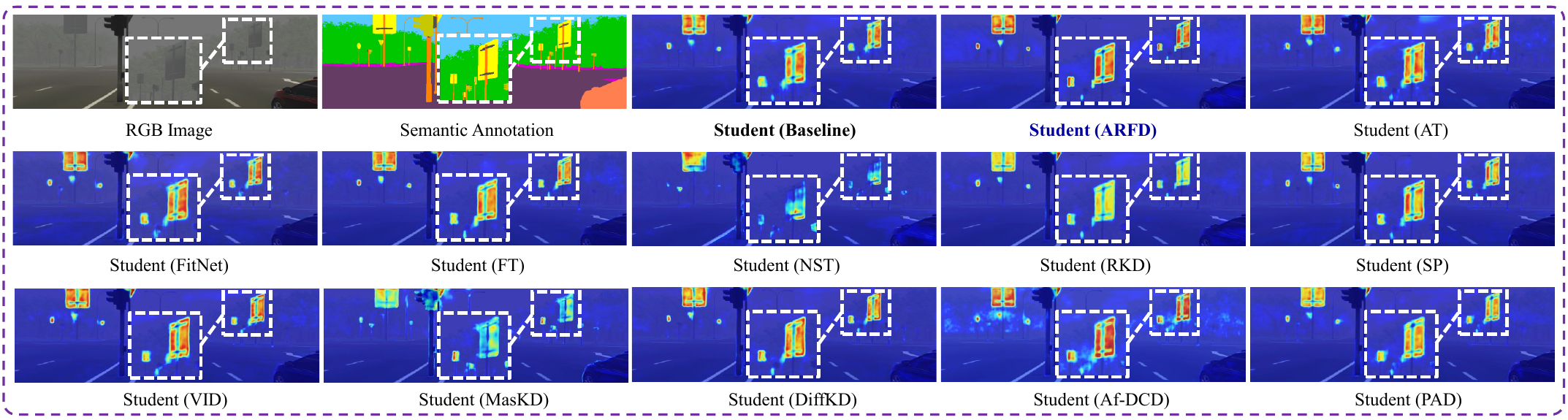}
		\centering
		\caption{\clr{Probabilities of the ``traffic sign'' class produced by student networks, where the red and blue colors correspond to the high and low probability of the predictions, respectively. Significantly improved areas are highlighted with white dashed boxes.}
}
		\label{fig.featuremap}
\end{figure*}

\clr{We first compare DWLD with the baseline algorithm DKD \cite{zhao2022decoupled}, which requires a manually-set, fixed $\beta$.} As shown in Fig. \ref{fig.ld}(a), DWLD consistently demonstrates superior performance over DKD across different values of $\beta$. These results suggest that, compared to DKD, which sometimes struggles to find a proper single, fixed weight $\beta$, our proposed DWLD offers a preferable option for both achieving better performance and simplifying the deployment process. 

\clr{Moreover, we validate the effectiveness of DWC both when $\boldsymbol{\Omega}^{\mathcal{S}}$ is set to $\hat{{\boldsymbol{P}}}^{\mathcal{S}}$ and when $\boldsymbol{c}^{\mathcal{S}}$ is additionally incorporated into $\boldsymbol{\Omega}^{\mathcal{S}}$. As depicted in Table \ref{tb.omega}, when $\boldsymbol{\Omega}^{\mathcal{S}}=\operatorname{Concat}\big(\hat{\boldsymbol{P}}^{\mathcal{S}}, \boldsymbol{c}^{\mathcal{S}}\big)$, the student network achieves increases by over 1.0\% in mIoU and by over 0.9\% in mFsc, supporting our claim regarding the design of $\boldsymbol{\Omega}^{\mathcal{S}}$.}

\clr{In theory, greater confidence in the teacher network should lead to a better distillation of non-target class information into the student network. While the design of the DWC draws inspiration from differentiating DKD \cite{zhao2022decoupled} concerning $z^{\mathcal{S}}_k$, there remains an open question whether the confidence of the teacher network also exerts a significant influence on DWC. Therefore, we further validate the effectiveness of DWC using both $\boldsymbol{\Omega}^{\mathcal{S}}$ and $\boldsymbol{\Omega}^{\mathcal{T}}$ as inputs, with and without the incorporation of dropout layers, as shown in Fig. \ref{fig.ld}(b).} As anticipated, DWC yields superior performance when utilizing $\boldsymbol{\Omega}^{\mathcal{S}}$ as input, confirming the fundamental practicability of the core concept underlying our DWC design. Furthermore, Fig. \ref{fig.ld}(b) also provides readers with the quantitative results on the selection of MLP layers, which is another key aspect of our DWC design. \clr{It is evident that as the number of MLP layers increases, the performance of the student network shows a gradual improvement until reaching a saturation point, after which its performance degrades due to over-fitting.} Additionally, the inclusion of dropout layers does not appear to enhance the overall performance of DWC. Therefore, our DWC utilizes three MLP layers without dropout.

\subsection{Ablation Study on Feature Distillation}
\label{sec.fd_ablation}

\begin{table}[!t]
\begin{center}
\settablefont
\caption{Comparisons of Various Feature Recalibration Methods and Feature Consistency Measurement Methods on the vKITTI2 dataset.}
\label{tb.feature}
\begin{tabular}{l|cccc}
\toprule
\diagbox{\makecell[c]{Feature \\Recalibration}}{\makecell[c]{Feature \\Consistency}} & \makecell[c]{Cosine \\Similarity} & \makecell[c]{Euclidean \\Distance}  & \makecell[c]{Pearson \\Correlation \\Coefficient}   &CKA  \\
\hline
\hline
Laplace-Based Kernel Regression&88.69  &90.64  &89.15  &\textbf{92.46}  \\
Gaussian-Based Kernel Regression&91.63  &91.26  &88.93  &91.33  \\
Linear-Based Kernel Regression&86.36  &90.50  &89.68  &91.25  \\
w/o Kernel Regression &86.39  &89.81  &88.51  &90.02  \\
\bottomrule
\end{tabular}
\end{center}
\end{table}

\clr{As presented in Table \ref{tb.feature}, remarkable improvements are achieved through feature recalibration via kernel regression.} These improvements can primarily be attributed to the effectiveness of kernel regression in reducing the gap between features in teacher and student networks across multiple dimensions. As expected, both Euclidean distance and the Pearson correlation coefficient prove effective for feature consistency measurement. However, when cosine similarity is used in conjunction with a linear kernel, it demonstrates even poorer performance, compared to the cases where no kernel regression is employed. We posit that conventional similarity measurement methods such as cosine similarity have the opposite effect by focusing on the differences in attribute values. \clr{Consequently, these methods are often misled by irrelevant details, such as image backgrounds, while overlooking the essential features.} In contrast, our ARFD, which leverages CKA based on the HSIC, offers a more comprehensive quantification of feature consistency. The optimal performance is achieved when combining Laplace-based kernel regression with CKA-based feature consistency measurement.

\clr{Furthermore, Fig. \ref{fig.cka} provides readers with a quantitative comparison of the CKA scores between the features of the baseline and distilled student networks. As for the distilled student network, the CKA score of the features steadily increases and stabilizes as training progresses. In contrast, the CKA score of the features from the baseline student network exhibits continuous fluctuations throughout the training process. This comparison highlights that, without feature consistency supervision, it is challenging for the single-model student network to achieve stable feature alignment with the teacher network. This further highlights the effectiveness of the ARFD in facilitating consistent knowledge transfer.}

\clr{We also provide a qualitative comparison between our proposed ARFD and other LD approaches. As depicted in Fig. \ref{fig.featuremap}, ARFD generates more confident predictions. Notably, it leads to a more uniform distribution of probabilities for the ``traffic sign'' class, indicating the effective infusion of spatial geometric prior knowledge through ARFD.}

\clr{\subsection{Additional Experiments}
\label{sec:add}
To comprehensively validate the effectiveness of our proposed LIX framework, we conduct several additional experiments, as detailed in the supplementary material. Specifically, we evaluate LIX's generalizability by training the student network on one dataset and testing it on another. Furthermore, we utilize Transformer-based OFF-Net \cite{min2022orfd} and real-world CityScapes dataset \cite{cordts2016cityscapes} to further validate the compatibility of our proposed LIX framework across different architectures and datasets. Additionally, we compare LIX's training overheads with other KD methods and conduct a per-category performance evaluation to quantify improvements across different categories. Furthermore, we provide visual comparisons of KD loss distribution, NCLD weight $\boldsymbol{\beta}^{\mathcal{S}}$, and the confidence map to enhance interpretability. These results provide additional quantitative and qualitative evaluations, highlighting LIX's superior performance, generalizability, and limitations. }

\section{Discussion}
\label{sec:discussion}
\begin{figure}[!t]
		\centering
		\includegraphics[width=0.49\textwidth]{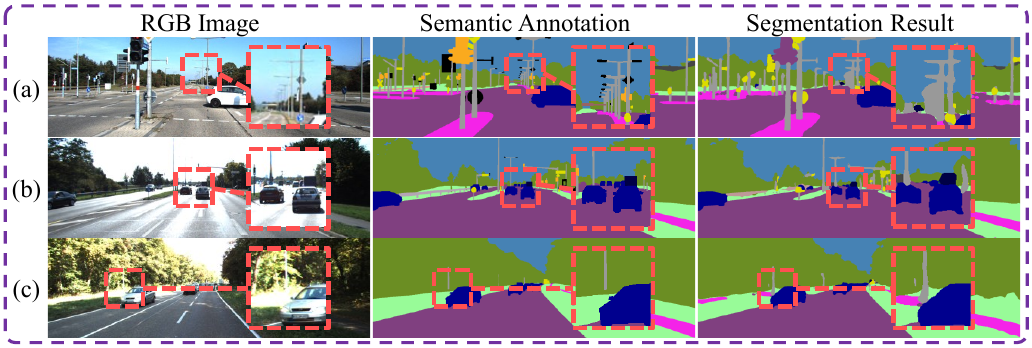}
		\centering
		\caption{\clr{Unsatisfactory results on the KITTI Semantics dataset: (a) congested objects; (b) distant areas; (c) over-exposed areas. Challenging areas are highlighted with red dashed boxes.}
            }
		\label{fig.challenge}
\end{figure}

\clr{While LIX demonstrates significant advancements, it faces challenges in certain scenarios and requires further improvements. As depicted in Fig. \ref{fig.challenge}(a)(c), when depth is inaccurately estimated, especially in areas with congested objects or over-exposure, the teacher network may hinder the effective transfer of spatial geometric knowledge. Additionally, as detailed in Fig. \ref{fig.challenge}(b), the segmentation of distant objects remains unsatisfactory due to ambiguous representations in low-resolution feature maps. Moreover, although our LIX framework improves the network's generalizability to some extent, the zero-shot performance of the student network remains suboptimal. This limitation may stem from significant variations in spatial geometric information across datasets, notably differences in depth ranges. Addressing these limitations presents promising directions for future research and optimization.}

\clr{Additionally, since the LIX framework is inherently modular, it is extendable to fields beyond autonomous driving. DWLD is applicable to any task involving softmax-based outputs, such as image classification and object detection. ARFD, benefiting from its feature recalibration and CKA-based consistency measurement, can be adapted to dense prediction tasks like medical image segmentation and remote sensing. As a versatile distillation framework, LIX's ability to handle diverse data modalities makes it a promising solution for applications such as multi-modal learning and cross-domain adaptation. For instance, in the crop monitoring task, LIX can infuse prior knowledge from a multi-spectral teacher network into a student network trained exclusively with RGB images. This flexibility highlights LIX's potential as a general framework for knowledge distillation across various fields.
}

\section{Conclusion and Future Work}
\label{sec:conclusions}
\clr{This article discussed a new computer vision problem: the implicit infusion of spatial geometric prior knowledge acquired by a data-fusion teacher network into a single-modal student network. We contributed to both logit distillation and feature distillation by introducing the DWLD and the ARFD algorithms, respectively. We extended the DKD algorithm, by introducing a logit-wise dynamic weight controller, which assigns an appropriate weight to each logit. As for FD, we introduced two novel techniques: feature recalibration via kernel regression and in-depth feature consistency quantification via CKA. Through extensive experiments conducted with representative RGB-X semantic segmentation networks on public autonomous driving datasets, we validated the effectiveness and superior performance of our developed LIX framework. Our future work will primarily concentrate on refining the designs of LIX for greater feasibility and generalizability.}

{
    \small
    \bibliographystyle{IEEEtran}
    \bibliography{main}

\begin{thebibliography}{10}
\providecommand{\url}[1]{#1}
\csname url@samestyle\endcsname
\providecommand{\newblock}{\relax}
\providecommand{\bibinfo}[2]{#2}
\providecommand{\BIBentrySTDinterwordspacing}{\spaceskip=0pt\relax}
\providecommand{\BIBentryALTinterwordstretchfactor}{4}
\providecommand{\BIBentryALTinterwordspacing}{\spaceskip=\fontdimen2\font plus
\BIBentryALTinterwordstretchfactor\fontdimen3\font minus \fontdimen4\font\relax}
\providecommand{\BIBforeignlanguage}[2]{{%
\expandafter\ifx\csname l@#1\endcsname\relax
\typeout{** WARNING: IEEEtran.bst: No hyphenation pattern has been}%
\typeout{** loaded for the language `#1'. Using the pattern for}%
\typeout{** the default language instead.}%
\else
\language=\csname l@#1\endcsname
\fi
#2}}
\providecommand{\BIBdecl}{\relax}
\BIBdecl

\bibitem{li2024roadformer}
J.~Li \emph{et~al.}, ``{RoadFormer}: Duplex {Transformer} for {RGB-Normal} semantic road scene parsing,'' \emph{IEEE Transactions on Intelligent Vehicles}, vol.~9, no.~7, pp. 5163--5172, 2024.

\bibitem{fan2020sne-roadseg}
R.~Fan \emph{et~al.}, ``{SNE-RoadSeg}: Incorporating surface normal information into semantic segmentation for accurate freespace detection,'' in \emph{Proceedings of the European Conference on Computer Vision (ECCV)}.\hskip 1em plus 0.5em minus 0.4em\relax Springer, 2020, pp. 340--356.

\bibitem{zhang2023cmx}
J.~Zhang \emph{et~al.}, ``{CMX}: Cross-modal fusion for {RGB-X} semantic segmentation with transformers,'' \emph{IEEE Transactions on Intelligent Transportation Systems}, 2023.

\bibitem{yin2023dformer}
\BIBentryALTinterwordspacing
B.~Yin \emph{et~al.}, ``{DFormer}: Rethinking {RGBD} representation learning for semantic segmentation,'' \emph{Computing Research Repository {(CoRR)}}, vol. abs/2309.09668, 2023. [Online]. Available: \url{https://arxiv.org/abs/2309.09668}
\BIBentrySTDinterwordspacing

\bibitem{zhang2023delivering}
J.~Zhang \emph{et~al.}, ``Delivering arbitrary-modal semantic segmentation,'' in \emph{Proceedings of the IEEE/CVF Conference on Computer Vision and Pattern Recognition (CVPR)}, 2023, pp. 1136--1147.

\bibitem{wu2024s}
Z.~Wu \emph{et~al.}, ``S$^3${M-Net}: Joint learning of semantic segmentation and stereo matching for autonomous driving,'' \emph{IEEE Transactions on Intelligent Vehicles}, vol.~9, no.~2, pp. 3940--3951, 2024.

\bibitem{10292885}
C.~{Li} \emph{et~al.}, ``Boosting knowledge distillation via intra-class logit distribution smoothing,'' \emph{IEEE Transactions on Circuits and Systems for Video Technology}, vol.~34, no.~6, pp. 4190--4201, 2024.

\bibitem{shu2021channel}
C.~Shu \emph{et~al.}, ``Channel-wise knowledge distillation for dense prediction,'' in \emph{Proceedings of the IEEE/CVF International Conference on Computer Vision (ICCV)}, 2021, pp. 5311--5320.

\bibitem{hinton2015distilling}
\BIBentryALTinterwordspacing
G.~Hinton \emph{et~al.}, ``Distilling the knowledge in a neural network,'' \emph{Computing Research Repository {(CoRR)}}, vol. abs/1503.02531, 2015. [Online]. Available: \url{https://arxiv.org/abs/1503.02531}
\BIBentrySTDinterwordspacing

\bibitem{zagoruyko2016paying}
\BIBentryALTinterwordspacing
S.~Zagoruyko and N.~Komodakis, ``Paying more attention to attention: Improving the performance of convolutional neural networks via attention transfer,'' \emph{Computing Research Repository {(CoRR)}}, vol. abs/1612.03928, 2016. [Online]. Available: \url{https://arxiv.org/abs/1612.03928}
\BIBentrySTDinterwordspacing

\bibitem{kim2018paraphrasing}
J.~Kim \emph{et~al.}, ``Paraphrasing complex network: Network compression via factor transfer,'' \emph{Advances in Neural Information Processing Systems (NeurIPS)}, vol.~31, 2018.

\bibitem{huang2017like}
\BIBentryALTinterwordspacing
Z.~Huang and N.~Wang, ``Like what you like: Knowledge distill via neuron selectivity transfer,'' \emph{Computing Research Repository {(CoRR)}}, vol. abs/1707.01219, 2017. [Online]. Available: \url{https://arxiv.org/abs/1707.01219}
\BIBentrySTDinterwordspacing

\bibitem{park2019relational}
W.~Park \emph{et~al.}, ``Relational knowledge distillation,'' in \emph{Proceedings of the IEEE/CVF Conference on Computer Vision and Pattern Recognition (CVPR)}, 2019, pp. 3967--3976.

\bibitem{tung2019similarity}
F.~Tung and G.~Mori, ``Similarity-preserving knowledge distillation,'' in \emph{Proceedings of the IEEE/CVF International Conference on Computer Vision (ICCV)}, 2019, pp. 1365--1374.

\bibitem{ahn2019variational}
S.~Ahn \emph{et~al.}, ``Variational information distillation for knowledge transfer,'' in \emph{Proceedings of the IEEE/CVF Conference on Computer Vision and Pattern Recognition (CVPR)}, 2019, pp. 9163--9171.

\bibitem{chen2021distilling}
P.~Chen \emph{et~al.}, ``Distilling knowledge via knowledge review,'' in \emph{Proceedings of the IEEE/CVF Conference on Computer Vision and Pattern Recognition (CVPR)}, 2021, pp. 5008--5017.

\bibitem{zheng2022boosting1}
W.~{Zheng} \emph{et~al.}, ``Boosting {3D} object detection by simulating multimodality on point clouds,'' in \emph{Proceedings of the IEEE/CVF Conference on Computer Vision and Pattern Recognition (CVPR)}, 2022, pp. 13\,638--13\,647.

\bibitem{zhou2023unidistill}
S.~Zhou \emph{et~al.}, ``{UniDistill}: A universal cross-modality knowledge distillation framework for {3D} object detection in bird's-eye view,'' in \emph{Proceedings of the IEEE/CVF Conference on Computer Vision and Pattern Recognition (CVPR)}, 2023, pp. 5116--5125.

\bibitem{cen2023cmdfusion}
J.~Cen \emph{et~al.}, ``{CMDFusion}: Bidirectional fusion network with cross-modality knowledge distillation for {LiDAR} semantic segmentation,'' \emph{IEEE Robotics and Automation Letters}, vol.~9, no.~1, pp. 771--778, 2023.

\bibitem{zheng2022boosting2}
W.~{{Zheng}} \emph{et~al.}, ``Boosting single-frame {3D} object detection by simulating multi-frame point clouds,'' in \emph{Proceedings of the 30th ACM International Conference on Multimedia ({ACM MM})}, 2022, pp. 4848--4856.

\bibitem{qiu2023multi}
S.~Qiu \emph{et~al.}, ``Multi-to-single knowledge distillation for point cloud semantic segmentation,'' in \emph{2023 IEEE International Conference on Robotics and Automation (ICRA)}.\hskip 1em plus 0.5em minus 0.4em\relax IEEE, 2023, pp. 9303--9309.

\bibitem{aggarwal2015data}
C.~C. Aggarwal \emph{et~al.}, \emph{Data mining: the textbook}.\hskip 1em plus 0.5em minus 0.4em\relax Springer, 2015, vol.~1.

\bibitem{zhao2022decoupled}
B.~Zhao \emph{et~al.}, ``Decoupled knowledge distillation,'' in \emph{Proceedings of the IEEE/CVF Conference on Computer Vision and Pattern Recognition (CVPR)}, 2022, pp. 11\,953--11\,962.

\bibitem{nguyen2020wide}
\BIBentryALTinterwordspacing
T.~Nguyen \emph{et~al.}, ``Do wide and deep networks learn the same things? uncovering how neural network representations vary with width and depth,'' \emph{Computing Research Repository {(CoRR)}}, vol. abs/2010.15327, 2020. [Online]. Available: \url{https://arxiv.org/abs/2010.15327}
\BIBentrySTDinterwordspacing

\bibitem{ma2020hsic}
W.-D.~K. Ma \emph{et~al.}, ``The {HSIC} bottleneck: Deep learning without back-propagation,'' in \emph{Proceedings of the AAAI Conference on Artificial Intelligence (AAAI)}, vol.~34, no.~04, 2020, pp. 5085--5092.

\bibitem{zhang2021deep}
Y.~Zhang \emph{et~al.}, ``Deep multimodal fusion for semantic image segmentation: A survey,'' \emph{Image and Vision Computing}, vol. 105, p. 104042, 2021.

\bibitem{wang2021sne}
H.~Wang \emph{et~al.}, ``{SNE-RoadSeg+}: Rethinking depth-normal translation and deep supervision for freespace detection,'' in \emph{2021 IEEE/RSJ International Conference on Intelligent Robots and Systems (IROS)}.\hskip 1em plus 0.5em minus 0.4em\relax IEEE, 2021, pp. 1140--1145.

\bibitem{ha2017mfnet}
Q.~Ha \emph{et~al.}, ``{MFNet}: Towards real-time semantic segmentation for autonomous vehicles with multi-spectral scenes,'' in \emph{2017 IEEE/RSJ International Conference on Intelligent Robots and Systems (IROS)}.\hskip 1em plus 0.5em minus 0.4em\relax IEEE, 2017, pp. 5108--5115.

\bibitem{valada2017adapnet}
A.~Valada \emph{et~al.}, ``{AdapNet}: Adaptive semantic segmentation in adverse environmental conditions,'' in \emph{2017 IEEE International Conference on Robotics and Automation (ICRA)}.\hskip 1em plus 0.5em minus 0.4em\relax IEEE, 2017, pp. 4644--4651.

\bibitem{cheng2017locality}
Y.~Cheng \emph{et~al.}, ``Locality-sensitive deconvolution networks with gated fusion for {RGB-D} indoor semantic segmentation,'' in \emph{Proceedings of the IEEE Conference on Computer Vision and Pattern Recognition (CVPR)}, 2017, pp. 3029--3037.

\bibitem{10024393}
C.~Li \emph{et~al.}, ``Instance-aware distillation for efficient object detection in remote sensing images,'' \emph{IEEE Transactions on Geoscience and Remote Sensing}, vol.~61, pp. 1--11, 2023.

\bibitem{bang2024radardistill}
G.~{Bang} \emph{et~al.}, ``{RadarDistill}: Boosting {Radar-based }object detection performance via knowledge distillation from {LiDAR} features,'' in \emph{Proceedings of the IEEE/CVF Conference on Computer Vision and Pattern Recognition (CVPR)}, 2024, pp. 15\,491--15\,500.

\bibitem{zhou2024mstnet}
W.~Zhou \emph{et~al.}, ``{MSTNet-KD}: Multilevel transfer networks using knowledge distillation for the dense prediction of remote-sensing images,'' \emph{IEEE Transactions on Geoscience and Remote Sensing}, 2024.

\bibitem{zhourethinking}
H.~Zhou \emph{et~al.}, ``Rethinking soft labels for knowledge distillation: A bias--variance tradeoff perspective,'' in \emph{International Conference on Learning Representations {(ICLR)}}, 2021.

\bibitem{huang2022knowledge}
T.~Huang \emph{et~al.}, ``Knowledge distillation from a stronger teacher,'' \emph{Advances in Neural Information Processing Systems (NeurIPS)}, vol.~35, pp. 33\,716--33\,727, 2022.

\bibitem{liu2015representation}
X.~Liu \emph{et~al.}, ``Representation learning using multi-task deep neural networks for semantic classification and information retrieval,'' in \emph{Proceedings of the 2015 Conference of the North American Chapter of the Association for Computational Linguistics: Human Language Technologies (NAACL)}, 2015, pp. 912--921.

\bibitem{romero2014FitNets}
\BIBentryALTinterwordspacing
A.~Romero \emph{et~al.}, ``{FitNets}: Hints for thin deep nets,'' \emph{Computing Research Repository {(CoRR)}}, vol. abs/1412.6550, 2014. [Online]. Available: \url{https://arxiv.org/abs/1412.6550}
\BIBentrySTDinterwordspacing

\bibitem{huangmasked}
T.~{Huang} \emph{et~al.}, ``Masked distillation with receptive tokens,'' in \emph{International Conference on Learning Representations {(ICLR)}}, 2023.

\bibitem{fan2023augmentation}
J.~Fan \emph{et~al.}, ``Augmentation-free dense contrastive knowledge distillation for efficient semantic segmentation,'' \emph{Advances in Neural Information Processing Systems (NeurIPS)}, pp. 51\,359--51\,370, 2023.

\bibitem{zhang2020prime}
Y.~Zhang \emph{et~al.}, ``Prime-aware adaptive distillation,'' in \emph{Proceedings of the European Conference on Computer Vision (ECCV)}.\hskip 1em plus 0.5em minus 0.4em\relax Springer, 2020, pp. 658--674.

\bibitem{guo2023boosting}
Z.~{Guo} \emph{et~al.}, ``Boosting graph neural networks via adaptive knowledge distillation,'' in \emph{Proceedings of the AAAI Conference on Artificial Intelligence (AAAI)}, vol.~37, no.~6, 2023, pp. 7793--7801.

\bibitem{peng2019correlation}
B.~Peng \emph{et~al.}, ``Correlation congruence for knowledge distillation,'' in \emph{Proceedings of the IEEE/CVF International Conference on Computer Vision (ICCV)}, 2019, pp. 5007--5016.

\bibitem{heo2019knowledge}
B.~Heo \emph{et~al.}, ``Knowledge transfer via distillation of activation boundaries formed by hidden neurons,'' in \emph{Proceedings of the AAAI Conference on Artificial Intelligence (AAAI)}, vol.~33, no.~01, 2019, pp. 3779--3787.

\bibitem{yim2017gift}
J.~Yim \emph{et~al.}, ``A gift from knowledge distillation: Fast optimization, network minimization and transfer learning,'' in \emph{Proceedings of the IEEE Conference on Computer Vision and Pattern Recognition (CVPR)}, 2017, pp. 4133--4141.

\bibitem{kornblith2019similarity}
S.~Kornblith \emph{et~al.}, ``Similarity of neural network representations revisited,'' in \emph{International Conference on Machine Learning (ICML)}.\hskip 1em plus 0.5em minus 0.4em\relax PMLR, 2019, pp. 3519--3529.

\bibitem{cortes2012algorithms}
C.~Cortes \emph{et~al.}, ``Algorithms for learning kernels based on centered alignment,'' \emph{The Journal of Machine Learning Research}, vol.~13, no.~1, pp. 795--828, 2012.

\bibitem{menze2015kitti}
M.~Menze and A.~Geiger, ``Object scene flow for autonomous vehicles,'' in \emph{Proceedings of the IEEE/CVF Conference on Computer Vision and Pattern Recognition (CVPR)}, 2015, pp. 3061--3070.

\bibitem{nuscenes2019}
H.~Caesar \emph{et~al.}, ``{nuScenes}: A multimodal dataset for autonomous driving,'' in \emph{Proceedings of the IEEE/CVF Conference on Computer Vision and Pattern Recognition (CVPR)}, 2020, pp. 11\,621--11\,631.

\bibitem{cabon2020vkitti2}
\BIBentryALTinterwordspacing
Y.~Cabon \emph{et~al.}, ``{Virtual KITTI 2},'' \emph{Computing Research Repository {(CoRR)}}, vol. abs/2001.10773, 2020. [Online]. Available: \url{https://arxiv.org/abs/2001.10773}
\BIBentrySTDinterwordspacing

\bibitem{cordts2016cityscapes}
M.~Cordts \emph{et~al.}, ``The {CityScapes} dataset for semantic urban scene understanding,'' in \emph{Proceedings of the IEEE Conference on Computer Vision and Pattern Recognition (CVPR)}, 2016, pp. 3213--3223.

\bibitem{li2022practical}
J.~Li \emph{et~al.}, ``Practical stereo matching via cascaded recurrent network with adaptive correlation,'' in \emph{Proceedings of the IEEE/CVF Conference on Computer Vision and Pattern Recognition (CVPR)}, 2022, pp. 16\,263--16\,272.

\bibitem{yang2024depth}
L.~Yang \emph{et~al.}, ``{Depth Anything}: Unleashing the power of large-scale unlabeled data,'' in \emph{2024 IEEE/CVF Conference on Computer Vision and Pattern Recognition (CVPR)}, 2024, pp. 10\,371--10\,381.

\bibitem{huang2023knowledge}
T.~{{Huang}} \emph{et~al.}, ``Knowledge diffusion for distillation,'' \emph{Advances in Neural Information Processing Systems (NeurIPS)}, vol.~36, pp. 65\,299--65\,316, 2023.

\bibitem{robbins1951stochastic}
H.~Robbins and S.~Monro, ``A stochastic approximation method,'' \emph{The Annals of Mathematical Statistics}, pp. 400--407, 1951.

\bibitem{min2022orfd}
C.~Min \emph{et~al.}, ``{ORFD}: A dataset and benchmark for {OFF-road} freespace detection,'' in \emph{2022 International Conference on Robotics and Automation (ICRA)}.\hskip 1em plus 0.5em minus 0.4em\relax IEEE, 2022, pp. 2532--2538.

\end{thebibliography}
}

\setlength{\parindent}{0pt}
\footnotesize
\end{document}